\documentclass[10pt,twocolumn,letterpaper]{article}

\usepackage{cvpr}              
\usepackage{booktabs}
\usepackage{multirow}
\usepackage{hhline}
\usepackage{makecell}
\usepackage{graphicx}

\usepackage[dvipsnames]{xcolor}

\usepackage{amsmath,amsfonts,bm}

\def\eqref#1{equation~\ref{#1}}

\def\1{\bm{1}}

\def\mA{{\bm{A}}}

\def\mD{{\bm{D}}}

\def\mI{{\bm{I}}}

\def\mK{{\bm{K}}}

\def\mQ{{\bm{Q}}}

\def\mU{{\bm{U}}}
\def\mV{{\bm{V}}}

\def\mX{{\bm{X}}}

\DeclareMathAlphabet{\mathsfit}{\encodingdefault}{\sfdefault}{m}{sl}
\SetMathAlphabet{\mathsfit}{bold}{\encodingdefault}{\sfdefault}{bx}{n}

\usepackage{times}
\usepackage{epsfig}
\usepackage{graphicx}
\usepackage{amsmath}
\usepackage{amssymb}
\usepackage{float}
\usepackage{tabularx}
\usepackage{rotating}

\def\eg{\emph{e.g.,}}           
\def\etal{\emph{et al.}}                             

\definecolor{cvprblue}{rgb}{0.21,0.49,0.74}
\usepackage[breaklinks,colorlinks,citecolor=cvprblue]{hyperref}
\def\paperID{} 
\def\confName{CVPR}
\def\confYear{2024}

\title{Deciphering ‘What’ and ‘Where’ Visual Pathways from\\Spectral Clustering of Layer-Distributed Neural Representations}
\author{Xiao Zhang$^*$\\
University of Chicago\\
{\tt\small zhang7@uchicago.edu}
\and
David Yunis$^*$\\
TTI-Chicago\\
{\tt\small dyunis@ttic.edu}
\and
Michael Maire\\
University of Chicago\\
{\tt\small mmaire@uchicago.edu}
}

\newcommand{\HLINE}{\Xhline{4\arrayrulewidth}}

\begin{document}
\twocolumn[{%
   \renewcommand\twocolumn[1][]{#1}%
   \vspace{-20pt}
   \maketitle%
   \begin{minipage}[t]{1.0\linewidth}%
      \captionsetup{type=figure}%
      \vspace{-10pt}
      \begin{center}
         \begin{minipage}[t]{0.34\linewidth}%
            \vspace{0pt}
            \begin{center}
               \begin{minipage}[t]{0.05\linewidth}%
                  \vspace{11pt}
                  \begin{sideways}\scriptsize{\textbf{\textsf{Regions~~~Eigs~~~~~Image~~}}}\end{sideways}
               \end{minipage}%
               \begin{minipage}[t]{0.94\linewidth}%
                  \vspace{0pt}
                  \begin{center}
                     \scriptsize{\textbf{\textsf{Per-Image Grouping}}}\\
                     \vspace{-4pt}
                     \textcolor{gray}{\rule{\linewidth}{0.75pt}}
                  \end{center}
                  \begin{center}
                     \vspace{-9pt}
                     \includegraphics[width=1.0\linewidth]{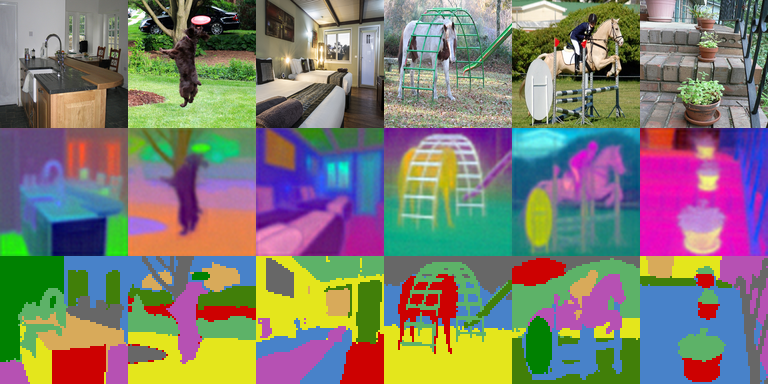}
                  \end{center}
               \end{minipage}%
            \end{center}
         \end{minipage}%
         \hfill
         \begin{minipage}[t]{0.005\linewidth}%
            \vspace{10pt}
            \begin{sideways}\textcolor{gray}{\rule{79.5pt}{0.75pt}}\end{sideways}
         \end{minipage}%
         \hfill
         \begin{minipage}[t]{0.64566\linewidth}%
            \vspace{0pt}
            \begin{center}
               \scriptsize{\textbf{\textsf{Full Dataset Grouping with Intra- and Inter-Image Affinity}}}\\
               \vspace{-4pt}
               \textcolor{gray}{\rule{\linewidth}{0.75pt}}
            \end{center}
            \vspace{-30pt}
            \begin{center}
               \begin{minipage}[t]{0.495\linewidth}%
                  \vspace{0pt}
                  \begin{center}
                     \includegraphics[width=1.0\linewidth]{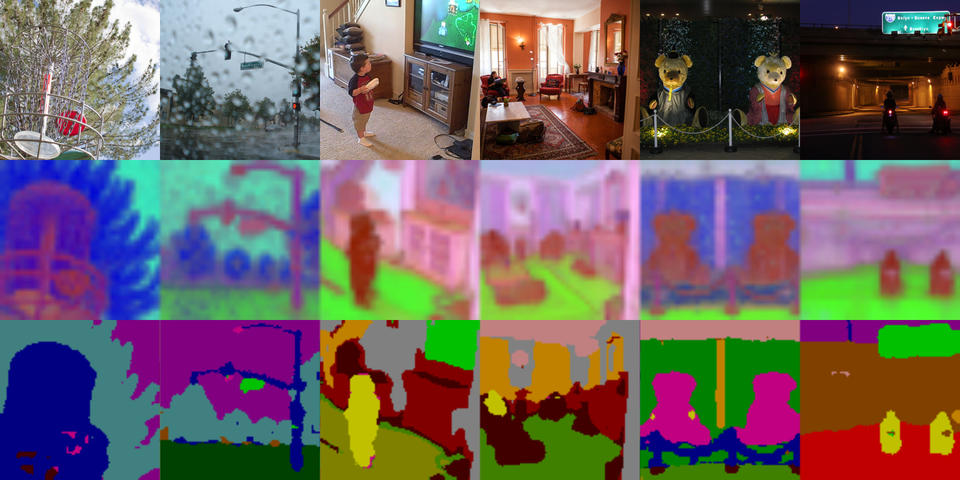}\\
                     \vspace{2pt}
                     \scriptsize{\textbf{\textsf{`What' Pathway}}}
                  \end{center}
               \end{minipage}%
               \hfill
               \begin{minipage}[t]{0.495\linewidth}%
                  \vspace{0pt}
                  \begin{center}
                     \includegraphics[width=1.0\linewidth]{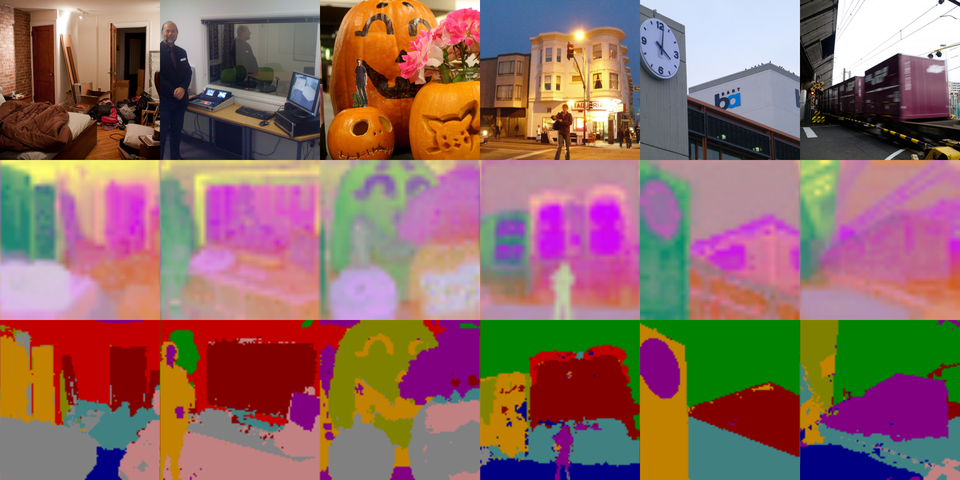}\\
                     \vspace{2pt}
                     \scriptsize{\textbf{\textsf{`Where' Pathway}}}
                  \end{center}
               \end{minipage}%
            \end{center}
         \end{minipage}%
      \end{center}
      \vspace{-15pt}
      \caption{
         Our novel optimization procedure, resembling spectral clustering, leverages features throughout layers of a
         pre-trained model to extract dense structural representations of images.  Shown are results of applying our
         method to Stable Diffusion~\cite{rombach2022high}.
         \emph{\textbf{Left:}}
         Analyzing internal feature affinity for a single input image yields region grouping.
         \emph{\textbf{Right:}}
         Extending the affinity graph across images yields coherent dataset-level segmentation and reveals `what'
         (object identity) and `where' (spatial location) pathways, depending on the feature source.
      }%
      \label{fig:teaser}
   \end{minipage}%
   \vspace{15pt}
}]%

\def\thefootnote{*}\footnotetext{Equal contribution. Code available at \href{https://github.com/xiao7199/layer_distributed_spectral_clustering}{this link}.} \def\thefootnote{\arabic{footnote}}
\begin{abstract}%
We present an approach for analyzing grouping information contained within a neural network's activations, permitting extraction of spatial layout and semantic segmentation from the behavior of large pre-trained vision models.  Unlike prior work, our method conducts a wholistic analysis of a network's activation state, leveraging features from all layers and obviating the need to guess which part of the model contains relevant information.  Motivated by classic spectral clustering, we formulate this analysis in terms of an optimization objective involving a set of affinity matrices, each formed by comparing features within a different layer.  Solving this optimization problem using gradient descent allows our technique to scale from single images to dataset-level analysis, including, in the latter, both intra- and inter-image relationships.  Analyzing a pre-trained generative transformer provides insight into the computational strategy learned by such models.  Equating affinity with key-query similarity across attention layers yields eigenvectors encoding scene spatial layout, whereas defining affinity by value vector similarity yields eigenvectors encoding object identity.  This result suggests that key and query vectors coordinate attentional information flow according to spatial proximity (a `where' pathway), while value vectors refine a semantic category representation (a `what' pathway).
\end{abstract}%

\section{Introduction}
\label{sec:intro}

\begin{figure*}[t]
   \begin{center}
      \begin{minipage}{1.0\linewidth}
      \begin{center}
         \includegraphics[width=1.0\linewidth]{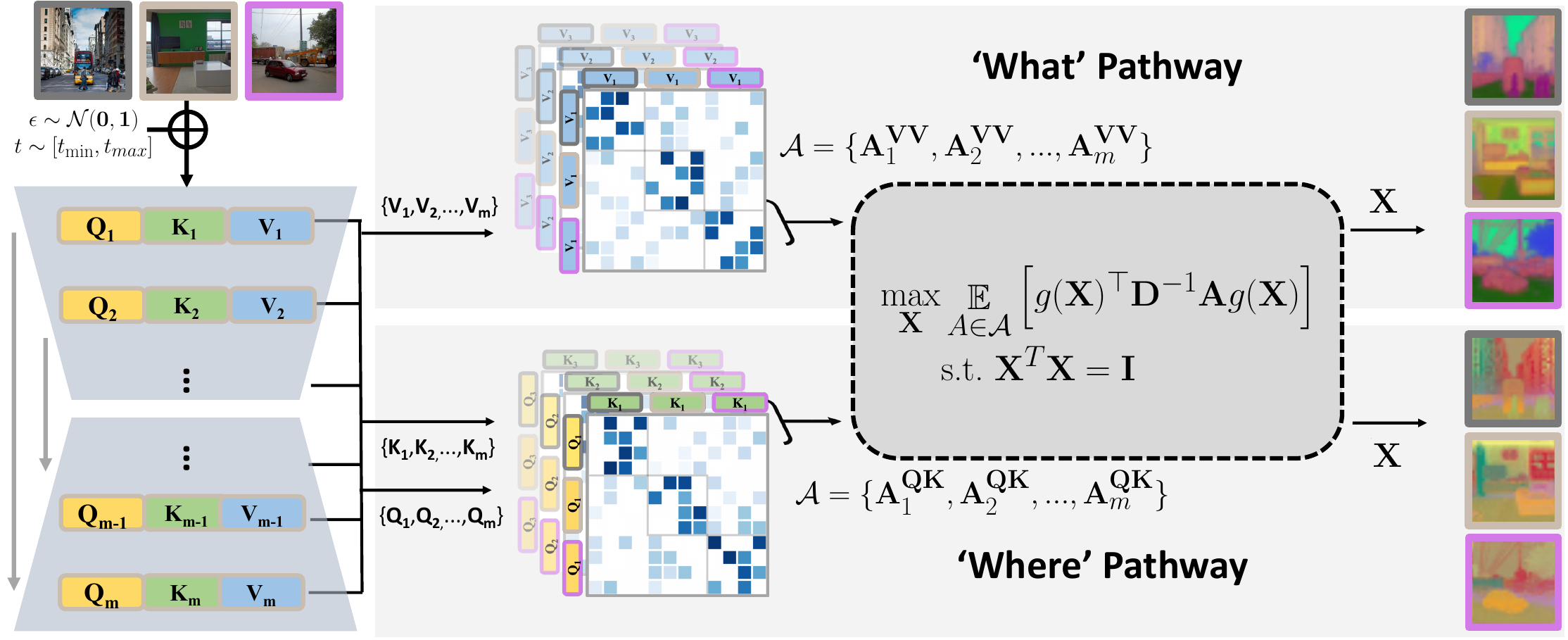}
         \vspace{-15pt}
         \caption{%
            \textbf{Spectral clustering of layer-distributed representations.}
            For each input image, we collect key, query, and value feature vectors from attention layers across
            network depth (and, for diffusion models, time).  Intra- and inter-image value-value (top) and key-query
            (bottom) similarity define a collection of affinity matrices indexed by layer (and time).  We solve for
            pseudo-eigenvectors $\mX$ which, when scaled to the spatial resolution of each layer via $g(\cdot)$, best
            satisfy an average of per-layer spectral partitioning criteria.  The leading eigenvector from value-value
            affinity reveals semantic category \emph{(top)}, while that from key-query affinity reveals spatial layout
            \emph{(bottom)}.%
         }%
         \label{fig:method}
      \end{center}
      \end{minipage}
   \end{center}
\vspace{-10pt}
\end{figure*}

An explosion in self-supervised learning techniques, including
adversarial~\cite{goodfellow2014generative,karras2019style,kang2023scaling},
contrastive~\cite{wu2018unsupervised,he2020momentum,chen2020improved,chen2020simple},
reconstructive~\cite{kingma2013auto,van2017neural}, and
denoising~\cite{sohl2015deep,ho2020denoising}
approaches, combined with the focus on training large-scale
foundation models~\cite{bommasani2021opportunities}
on vast collections of image data has produced deep neural networks exhibiting dramatic new capabilities.
Recent examples of such models include
CLIP~\cite{radford2021learning},
DINO~\cite{caron2021emerging},
MAE~\cite{he2022masked},
and Stable Diffusion~\cite{rombach2022high}.
As training is no longer primarily driven by annotated data, there is a critical need to understand what these models
have learned, provide interpretable insight into how they work, and develop techniques for porting their learned
representations for use in accomplishing additional tasks.

However, interpretable analysis of neural networks is challenging.  Procedures such as guided backpropagation~\cite{
springenberg2014striving} or Grad-CAM~\cite{selvaraju2017grad} assist with interpretability with respect to
particular labels, but are limited in scope.  Others propose heuristics for extracting information relevant to
particular downstream tasks, or analyze specific features in models~\cite{
dombrowski2022zero,
li2022adapting,
lis2022attentropy,
zhou2022extract,
tang2023emergent,
chen2023beyond,
hedlin2023unsupervised,
chefer2021generic,
chefer2021transformer,
baranchuk2021label}.
The distributed nature of both the information encoded within deep networks~\cite{szegedy2013intriguing} and their
computational structure frustrates the development of general-purpose techniques.

It is similarly unclear how best to repurpose pre-trained models toward downstream tasks.  Task-specific heuristics,
fine-tuning on labeled data, prompt engineering (if applicable), or clustering frozen feature representations might
all be viable options.  Yet, an element of art remains in choosing which features to extract or which layers to
fine-tune.

We introduce a new analysis approach that provides insight into model function and directly extracts significant
visual information about image segmentation, as shown in Figure~\ref{fig:teaser}, with neither a-priori knowledge of,
nor hyperparameter search over, where such information is stored in the network.  We accomplish this through an
analysis that couples the entire activation state of the network, from shallow to deep layers, into a global
spectral clustering objective.  Solving this clustering problem not only yields new feature representations (in the
form of eigenvector embeddings) directly relevant to downstream segmentation tasks, but also, as
Figure~\ref{fig:method} illustrates, provides insight into the inner workings of vision models.
Our contributions include:
\begin{itemize}[leftmargin=0.15in]
   \item{%
      A new approach, inspired by spectral clustering, for wholistic analysis of deep neural network activations.%
   }%
   \item{%
      Improved quality of extracted regions across models, compared to variants analyzing single layers.%
   }%
   \item{%
      An efficient gradient-based optimization framework that enables our approach to scale to joint analysis of
      network behavior across an entire dataset simultaneously.%
   }%
   \item{%
      Unsupervised semantic segmentation results on par with STEGO~\cite{hamilton2022unsupervised}, but
      extracted from a pre-trained generative model rather than a contrastive backbone.%
   }%
   \item{%
      Insight into the computational strategy learned by large-scale vision models: internal features are
      partitioned into `what' and `where' pathways, which separately maintain semantic and spatial information.%
   }%
\end{itemize}

\section{Related Work}

\noindent
\textbf{Image segmentation.}
Segmentation, as a generic grouping process, has historically been regarded as an important intermediate task in
computer vision.  Significant efforts focus on building object-agnostic methods for partitioning an image into
coherent regions or, equivalently, their dual representation as contours~\cite{Canny:PAMI:1986,AMFM:PAMI:2011,
RB:NIPS:2012,DZ:PAMI:2015,bertasius2015high,kokkinos2015pushing}, with standard benchmarks~\cite{MartinFTM01} driving
progress.  Semantic and instance segmentation, which aim to extract image regions corresponding to specific category
labels or object instances, have undergone parallel development, driven by benchmark datasets such as
PASCAL~\cite{everingham2015pascal} and COCO~\cite{lin2014microsoft}.  Notable modern supervised methods utilize
CNN~\cite{he2017mask} or Transformer~\cite{carion2020end} architectures trained in an end-to-end fashion.  Particularly
relevant is recent work demonstrating the ability of models to learn to segment with relatively few labels~\cite{
baranchuk2021label,ziegler2022self}.  Spectral clustering, as a method of approximating the solution of a graph
partitioning objective~\cite{shi2000normalized}, has appeared as a core algorithmic component across a variety of
segmentation systems~\cite{shi2000normalized,YGS:NIPS:2002,yu2004segmentation,AMFM:PAMI:2011,MYP:ICCV:2011,
kokkinos2015pushing,maire2016affinity,tang2018normalized}.

\vspace{0.25em}
\noindent
\textbf{Segmentation without labels.}
Recent methods, such as DINO~\citep{caron2021emerging}, learn intra-image and inter-image correspondences between
pixels without the need for dense labels.  STEGO~\cite{hamilton2022unsupervised} and PiCIE~\cite{cho2021picie} propose
to cluster pixel-wise features of a self-supervised backbone, showing impressive performance on semantic segmentation
with no labels at all.  LSeg~\cite{li2022language} and GroupViT~\cite{xu2022groupvit} modify CLIP~\cite{
radford2021learning} to enable zero-shot open-vocabulary semantic segmentation.

Another class of methods builds entirely on top of existing models, with no additional training~\cite{
dombrowski2022zero,li2022adapting,lis2022attentropy,zhou2022extract}.  Recent attempts at instance segmentation~\cite{
wang2022freesolo,wang2022tokencut} yield impressive results through heuristic decoding strategies based on the
structure of a particular model's features (\eg~the final layer of DINO~\cite{caron2021emerging}).  Other work, based
on Stable Diffusion~\cite{rombach2022high}, finds unsupervised dense correspondences using the text embedding space as
a shared anchor~\cite{hedlin2023unsupervised} or through careful choice of features~\cite{tang2023emergent}.

\vspace{0.25em}
\noindent
\textbf{Interpretability.}
Grad-CAM~\cite{selvaraju2017grad}, layer-wise relevance propagation~\cite{montavon2019layer}, and guided
backpropagation~\cite{springenberg2014striving} provide heuristics to visualize the responsibility of different input
spatial regions for predictions of a deep neural network.  Other approaches visualize attention matrices to find
salient input regions for NLP~\cite{clark2019does,kovaleva2019revealing} and vision~\citep{xu2015show,
chefer2021generic,chefer2021transformer} tasks.  Chen~\etal~\cite{chen2023beyond} find evidence of depth information
inside Stable Diffusion.  Yet, visualizing and interpreting neural network behavior remains a challenging problem
due to the distributed nature of the representations they learn~\cite{szegedy2013intriguing}.

\vspace{0.25em}
\noindent
\textbf{Spectral clustering of neural features.}
TokenCut~\cite{wang2022tokencut}, MaskCut~\cite{wang2023cut}, and DSM~\cite{Melas-Kyriazi_2022_CVPR} define affinity
graphs using final features of a pre-trained DINO~\cite{caron2021emerging} model, and use spectral clustering
to segment the original image.  We take inspiration from these approaches and utilize them as baselines for
experimental comparison.  Our methodology differs in being global and accounting for features throughout the
network, rather than restricted to one layer.

\vspace{0.25em}
\noindent
\textbf{Neuroscience perspectives on visual processing streams.}
Trevarthen~\cite{trevarthen1968two} and Schneider~\cite{schneider1969two} propose the concept of separate visual
processing pathways in the brain for localization (`where') and discrimination (`what').  Mishkin~\etal~\cite{
mishkin1983object} review evidence for this specialization of processing in the monkey, while subsequent
work examines specialized pathways in terms of perception and action~\cite{goodale1992separate}, as well as
spatial memory and navigation~\cite{kravitz2011new}.  While these ideas motivate our investigation into information
stored in the key, query, and value vectors distributed throughout a Transformer architecture, the question of
relevance (if any) to biological vision systems is beyond our scope.

\section{Method}

Our method closely resembles spectral clustering applied simultaneously across attention layers within a given neural
network.  The following sections detail our full method and discuss different graph construction choices for spectral
clustering, which respectively allow us to extract different kinds of information from source models.

\subsection{Spectral Clustering with Distributed Features}

Spectral clustering formulates the data grouping problem from the view of graph partitioning.  It uses the eigenvectors
of the normalized Laplacian matrix to partition the data into balanced subgraphs with minimal cost of breaking
edges~\citep{shi2000normalized}.  Specifically, with a symmetric affinity matrix $\mA\in \mathbb{R}^{N\times N}$,
where $N$ denotes the total number of data points and entries $\mA_{ij}\geq 0$ measure the similarity between data
samples with indices $i$ and $j$, we can embed the data into a lower dimensional representation
$\mX\in \mathbb{R}^{N\times C}$, where $C$ denotes the number of feature channels and $C\ll N$, as the solution of the
following generalized eigenproblem:
\begin{equation}
(\mD - \mA)\mX = \lambda \mD\mX.
\label{eqn:ncut}
\end{equation}
$\mD$ is the diagonal degree matrix of $\mA$ with diagonal entries $\mD_{ii} = \sum_{j}\mA_{ij}$, and $\mX, \lambda$
are eigenvectors and eigenvalues respectively.  We can then produce a discrete partition from $\mX$ through K-Means
clustering.

Though spectral clustering is a powerful tool for data analysis, its performance is highly dependent on the choice of
affinity matrix.  Recent works \citep{wang2022tokencut, wang2023cut, Melas-Kyriazi_2022_CVPR} apply spectral clustering
on an affinity matrix constructed from features in the last layer of DINO~\citep{caron2021emerging}, yielding strong
performance in segmentation tasks.  However, the choice of graph may not be clear when the desired information is
distributed across the layers of a neural network, or noise levels in diffusion models~\citep{baranchuk2021label}.
Therefore, we extend Eqn.~\ref{eqn:ncut} to allow for constructing $\mA$ using multiple sources of information.  A
classic approach to solve Eqn.~\ref{eqn:ncut} with a set of affinity matrices,
   $\mathcal{A} = \{\mA_1, \mA_2, \dots, \mA_m\}$,
is the constrained spectral clustering problem~\citep{cour2005spectral}.  It constructs a block diagonal affinity
matrix from this set:
\begin{equation}
   \mA_\mathcal{A} =
      \begin{bmatrix}
         \mA_{1} & & 0 \\
         & \ddots & \\
         0 & & \mA_{m}
      \end{bmatrix}
   \label{eqn:full_mat}
\end{equation}
and imposes additional cross-scale consistency constraints.  However, the size of this matrix, and the computational
expense of solving the resulting eigenproblem, can become intractable with increasing $|\mathcal{A}|$.  Instead of
solving the original eigenproblem in Eqn.~\ref{eqn:ncut}, we solve an approximation:
\begin{align}
   \label{eqn:eigen_optimization}
   \max_{\mX} & \mathop{\mathbb{E}}_{\mA\in \mathcal{A}} \left [ g(\mX)^\top \mD^{-1}_{\mA}\mA g(\mX) \right ], \nonumber\\
   & \text{s.t.} ~ \mX^\top \mX = \mI 
\end{align}
where $g(\cdot)$ corresponds to the resampling function that bilinearly interpolates the spatial resolution of $\mX$
to match the size of $\mA$, allowing affinity matrices to be constructed from feature maps with varying resolutions.
In Eqn.~\ref{eqn:eigen_optimization}, we follow \citet{meila2000learning} to solve the spectral clustering from the
random walk perspective, since the random walk matrix and the attention matrix have the same format and the same
eigenvectors.  This objective encodes a Rayleigh quotient optimization simultaneously across affinities in
$\mathcal{A}$, which avoids the intractable exact solution and can naturally scale with increasing $|\mathcal{A}|$.

Notice that $\mD^{-1}_{\mA}\mA$ is a random walk matrix with maximum eigenvalue of $1$.  For numerical stability,
we impose a constraint to ensure that the maximum value of the objective does not exceed $1$.  In addition, we replace
the strict orthogonality requirement with a soft Frobenius regularization term whose coefficient is $1$.  Consequently,
our final optimization objective is:
\begin{equation}\label{eqn:primary}
   \min_{\mX}
      \mathop{\mathbb{E}}_{\mA \in \mathcal{A}} | g(\mX)^\top \mD^{-1}_{\mA}\mA g(\mX) - 1 | +
      \lVert \mX^\top\mX - \mI \rVert_F.
\end{equation}
We parameterize $\mX$ as a learnable feature map and solve for it using gradient-based optimization.  In the following
subsections, we discuss the choice of affinity set $\mathcal{A}$ and how that choice affects the information we
extract.

\subsection{Per-Image Analysis}\label{sec:per_image_method}

Attention layers in vision models naturally consider patch-wise relationships when computing the attention matrix.
We can use this matrix as an affinity graph for spectral clustering, which allows investigating how a model groups
regions in an image internally, without imposing outside heuristics.  For the Vision Transformer~\citep{
dosovitskiy2020image} and U-Net~\citep{ronneberger2015u} variants that include a total of $m$ attention blocks, we
build an affinity set $\mathcal{A} = \{\mA^{\mQ\mK}_l\}_{l=1}^m$ across layers, where $\mA^{\mQ\mK}_l$ is the
pre-softmax self-attention matrix~\citep{vaswani2017attention} at layer $l$.
\begin{equation}
   \mA^{\mQ\mK}_l = \exp \left ( \frac{\mQ_l \mK_l^\top}{\sqrt{d_l}} \right ) \in \mathbb{R}^{N \times N},
\end{equation}
where $\mQ_l, \mK_l \in \mathbb{R}^{N \times d_l}$ are the query and key matrices for that layer respectively, and
$d_l$ is the embedding dimension.

\subsection{Full-Dataset Extension}\label{sec:dataset_extension}

We can extend the self-attention operation in a single image to affinity matrix construction across different images.
This allows probing how models relate different regions across different images using their internal computational
structure.  Specifically, we construct graphs similar to single-image self-attention matrices by computing normalized
pairwise dot products between queries at every position in one image, and keys at every position in another.  Scaling
to large datasets, we extract one set of features $\mX_i$ for each image with index $i$ in the dataset.  To do this,
we optimize a mini-batch of features:
\begin{equation}
   \mX_\text{batch} = \begin{bmatrix}
      \mX_j \\
      \vdots \\
      \mX_k
   \end{bmatrix} \in \mathbb{R}^{(N \cdot B) \times C},
\end{equation}
and construct graphs over that mini-batch:
\begin{equation}
   \mA^{\mQ\mK}_l =
   \begin{bmatrix}
      \widehat{\mQ}_{j,l} \\
      \vdots \\
      \widehat{\mQ}_{k,l}
   \end{bmatrix}
   \begin{bmatrix}
      \widehat{\mK}^\top_{j,l} \dots \widehat{\mK}^\top_{k,l}
   \end{bmatrix}
   \in \mathbb{R}^{(N \cdot B) \times (N \cdot B)},
\end{equation}
where $\widehat{\mQ}_{j, l}, \widehat{\mK}_{j, l} \in \mathbb{R}^{N \times d_l}$ represent the queries and keys for
image $j$ at layer $l$ normalized to unit-norm, and there are $B$ images in a mini-batch.  We normalize vectors as
calibrating magnitudes across images is not trivial.

Though we limit the graph to a mini-batch, it is still prohibitively expensive to store and optimize over.  Thus,
we sparsify the graph by only keeping the top $c_\text{intra}$ intra-image connections and the top $c_\text{inter}$
inter-image connections for each location.  In addition, we set all values below a threshold to $0$.  To investigate
what kind of information models mix across spatial locations, we consider a similar affinity set
$\mathcal{A} = \{\mA^{\mV\mV}_l\}_{l=1}^m$ built from the value matrices $\widehat{\mV}_{i, l}$.

With these approximate layer-wise graphs, we optimize the objective in Eqn.~\ref{eqn:primary} a small number of steps
per mini-batch, then sample a new mini-batch of images and continue.  Finally, this process discovers a consistent set
of dense features for a dataset.  A visualization of the entire method can be found in Figure~\ref{fig:method}.

\subsection{Recovering Orthogonal Representations}\label{sec:reorth}

Eqn.~\ref{eqn:primary} suggests an approximate formulation of the spectral clustering problem.  While this results in
a structured $\mX$, it fails to enforce an orthogonal representation capable of separating distinct features into
channels.  To overcome this, we orthogonalize $\mX$ by finding the eigenvectors $\mU$ of a small matrix
$\mX^\top \mX \in \mathbb{R}^{C \times C}$. This is similar to the reorthogonalization step in approximate eigensolvers;
\eg~lines 36-38 of Algorithm 2 in \citet{maire2013progressive}. The final representation is given by:
\begin{equation}
    \mX_\text{ortho} = \mX \mU.
\end{equation}

After extracting these final features, we create hard assignments using K-Means clustering.

\section{Experiments}

Leveraging our method, we investigate how models group image regions internally.
In Section~\ref{sec:per_image_results} we see how models associate locations within an image.
Section~\ref{sec:full_dataset_results} examines the same behavior across images and discovers a spatial/semantic
split depending on the choice of internal features used for grouping.  We evaluate this phenomenon quantitatively,
deriving a high quality training-free unsupervised semantic segmentation from Stable Diffusion~\citep{rombach2022high}
in Section~\ref{sec:what_pathway}, as well as providing stronger evidence for spatial information pathways in
Section~\ref{sec:where_pathway}.

\subsection{Per-Image Region Extraction}\label{sec:per_image_results}

\begin{figure}[t]
   \begin{center}
      \includegraphics[width=1.0\linewidth]{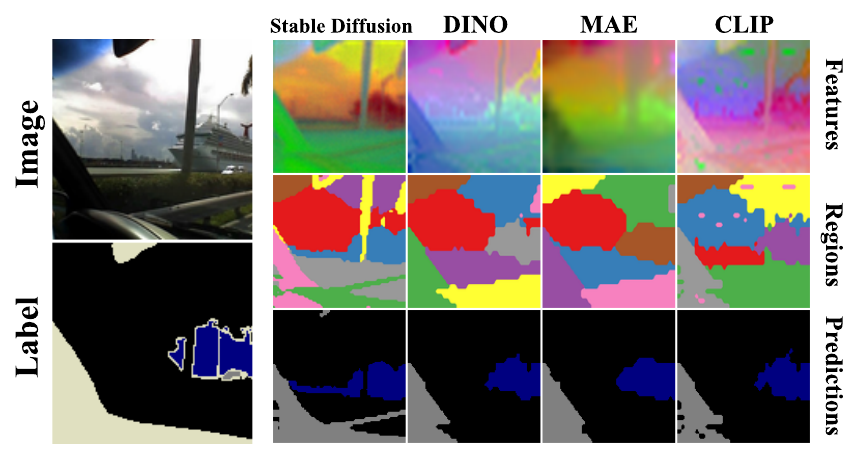}
   \end{center}
   \vspace{-1.0em}
   \caption{%
      \textbf{Features extracted from different models on PASCAL VOC~\citep{everingham2015pascal}.}
      Across models we extract meaningful regions, even for models like Stable Diffusion~\citep{
      rombach2022high}, CLIP~\citep{radford2021learning} or MAE~\citep{he2022masked} whose training is not
      well-aligned with segmentation.%
   }%
\label{fig:model_comparison}
\end{figure}

\begin{figure}[t]
   \vspace{1.0em}
   \begin{center}
      \includegraphics[width=1.0\linewidth]{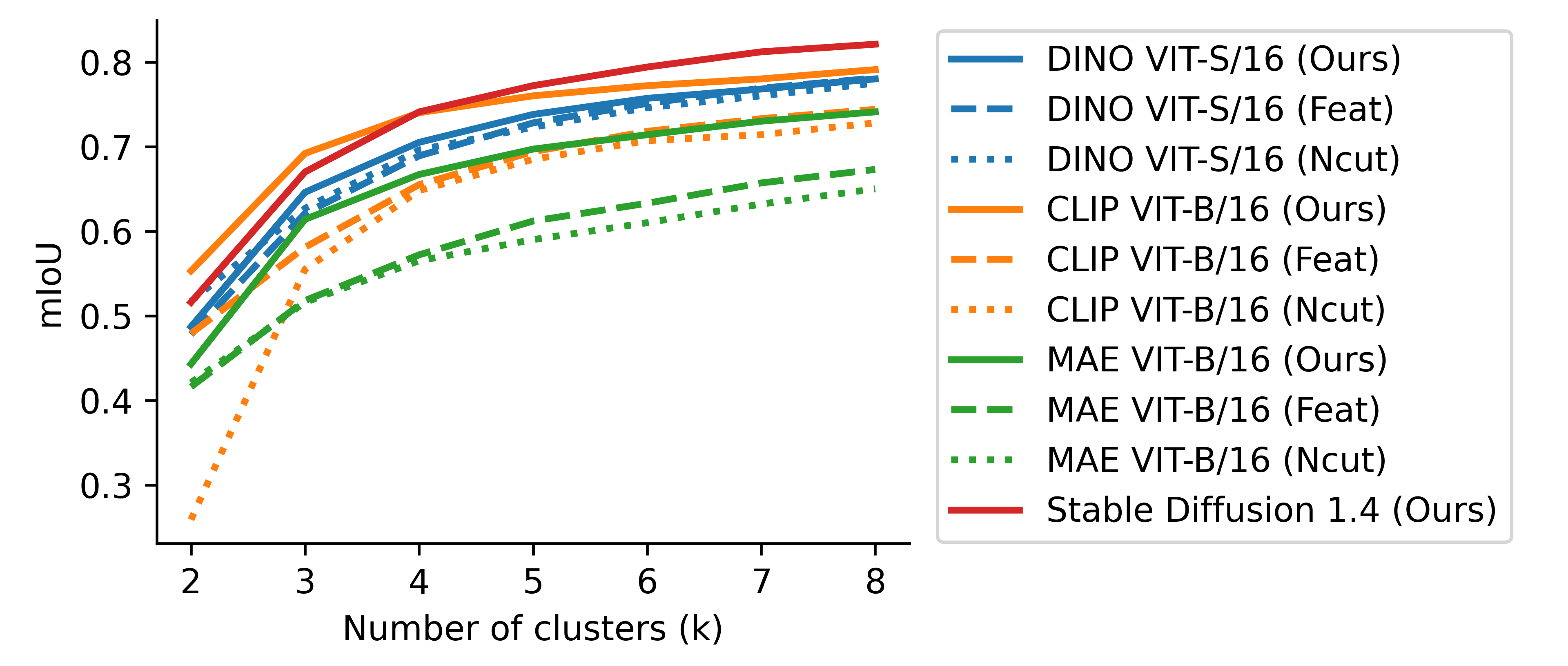}
   \end{center}
   \vspace{-1.0em}
   \caption{%
      \textbf{Oracle-based semantic segmentation performance with varying region count.}
      Across models and number of clusters (regions) returned by K-Means, our method (Ours + K-Means) yields better
      agreement (in mIoU) with ground-truth than running Normalized Cuts (Ncut + K-Means), or directly applying
      K-Means on the final output features of the model (K-Means).  We observe an even more significant improvement
      when applying our method to MAE and CLIP, which do not produce discriminative features.%
   }%
\label{fig:main_miou_vs_k}
\end{figure}

\begin{table}
\begin{center}
\footnotesize
\setlength{\tabcolsep}{5.25pt}
\begin{tabular}{@{}l|c|c|c@{}}
Model & Affinity Source & Mask & mIoU \\
\HLINE
Stable Diff.~1.4 \citep{rombach2022high} & All Attentions & Ours + K-Means & \textbf{0.82} \\
\hline
CLIP ViT-B/16 \citep{radford2021learning} & All Attentions & Ours + K-Means & \textbf{0.78} \\
CLIP ViT-B/16 \citep{radford2021learning} & Final Features & K-Means & 0.57 \\
CLIP ViT-B/16 \citep{radford2021learning} & Final Features & Ncut + K-Means \citep{wang2022tokencut} & 0.45 \\
\hline
DINO ViT-S/16 \citep{caron2021emerging} & All Attentions & Ours + K-Means & \textbf{0.78}\\
DINO ViT-S/16 \citep{caron2021emerging} & Final Attentions & Ncut + K-Means \citep{wang2022tokencut} & 0.58 \\
DINO ViT-S/16 \citep{caron2021emerging} & Final Features & K-Means & 0.74 \\
DINO ViT-S/16 \citep{caron2021emerging} & Final Features & Ncut + K-Means \citep{wang2022tokencut} & 0.73 \\
DINO ViT-S/16 \citep{caron2021emerging} & Final Features & MaskCut\citep{wang2023cut} &  0.64\\
\hline
MAE ViT-B/16 \citep{he2022masked} & All Attention & Ours + K-Means  & \textbf{0.74}\\
MAE ViT-B/16 \citep{he2022masked} & Final Features & Ncut + K-Means \citep{caron2021emerging}  & 0.62\\
MAE ViT-B/16 \citep{he2022masked} & Final Features & K-Means  & 0.48\\
\end{tabular}
\end{center}
\vspace{-1.0em}
\caption{%
   \textbf{Oracle decoding on PASCAL VOC \citep{everingham2015pascal} }.
   Compared with several strong baselines~\citep{wang2022tokencut,wang2023cut} applied to single-level features, our
   method can consistently extract accurate segmentation.  Our method works well even for models like CLIP~\citep{
   radford2021learning} and MAE~\citep{he2022masked}, whose final layer features are not discriminative enough for
   segmentation.  Our method is agnostic to the location of information, so we avoid this difficulty.
}
\label{tab:oracle_decoding}
\end{table}

To show how models partition images spatially, we extract dense eigenvector for individual images and
cluster these features into hard segmentations, as detailed in Section~\ref{sec:per_image_method}.

\begin{table}[!thp]
\begin{center}
\footnotesize 
\begin{subtable}{.5\textwidth}
\begin{tabular}{l|c|c|c|c|c|c}
\text{Config} & \textbf{All} & Enc & Mid & Dec & 32x32& 64x64\\
\HLINE

\text{Layer Index}&
 1-16 & 1-4 & 5 - 10 & 11-16 & \begin{tabular}{@{}c@{}} 3-4\\11-13\end{tabular} & \begin{tabular}{@{}c@{}} 1-2\\14-16\end{tabular}\\
 \hline
 mIoU & 0.75 & 0.66 & 0.76 & \textbf{0.80} & 0.75 & 0.70
\end{tabular}
\end{subtable}

\vspace{5pt}
\begin{subtable}{.5\textwidth}
\begin{tabular}{p{1.35cm}|p{0.515cm}|p{0.52cm}|p{0.64cm}|p{0.5cm}}
 $t_\text{max}$ & 250 & \textbf{500} & 750 & 999 \\
 \HLINE
 mIoU &\textbf{0.77} & 0.75 & 0.74 & 0.69
\end{tabular}
\end{subtable}
\vspace{-10pt}
\end{center}
\caption{Ablation of layer index and maximum noise level of the diffusion model on the PASCAL VOC dataset \citep{everingham2015pascal}. We find that using only decoder layers and middle noise yields the best results.}
\label{tab:ablate_layer_time}
\end{table}

\vspace{0.25em}
\noindent
\textbf{Experimental Setup.}
For all models, during optimization we consider all heads of all self-attention layers to be independent graphs.
In the case of Stable Diffusion, this is 16 self-attention layers with 8 attention heads, thus
$|\mathcal{A}| = 16 \times 8 = 128$ per forward pass.
Specific to Stable Diffusion, in each iteration we add noise to the input image by randomly sampling noise timestep
$t \in \mathcal{U}[0,500)$.  For all models, we construct feature map $\mX$ with spatial resolution matching the finest
attention layer resolution and set $C = 10$. We initialize $\mX$ from a normal distribution and solve the optimization
problem with Adam for $\sim 2000$ iterations with learning rate 1e-3.

To produce discrete regions, we run K-Means clustering by sweeping $K$ from 2 to 10 and use silhouette
score~\citep{rousseeuw1987silhouettes} to select the best value.  To speed up extraction in Stable Diffusion, we cache
attention matrices into a buffer for reuse with a 90$\%$ chance, bringing the per-image runtime from 154 to 67 seconds.
For more implementation details, please refer to Appendix~\ref{app:per_image_details}.  To provide a measure for
comparison, we extract multiple regions according to two related methods: Normalized Cut~\citep{shi2000normalized} and
MaskCut~\citep{wang2023cut}.  Both of these methods require a single affinity matrix, the choice of which we ablate in Appendix~\ref{app:per_image_details}.

\begin{figure}[t]
   \begin{center}
   \begin{minipage}[t]{0.03\linewidth}%
      \vspace{-30pt}
      \begin{sideways}\scriptsize{\textbf{\textsf{Image}}}\end{sideways}
   \end{minipage}%
   \begin{minipage}[t]{0.95\linewidth}%
      \begin{center}
         \includegraphics[width=1.0\linewidth]{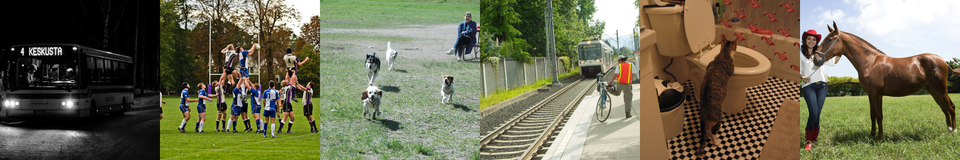}
      \end{center}
   \end{minipage}%

   \begin{minipage}[t]{0.03\linewidth}%
      \vspace{-68pt}
      \begin{sideways}\scriptsize{\textbf{\textsf{Eigs (Q-K Graph)}}}\end{sideways}
   \end{minipage}%
   \begin{minipage}[t]{0.95\linewidth}%
      \begin{center}
         \includegraphics[width=1.0\linewidth]{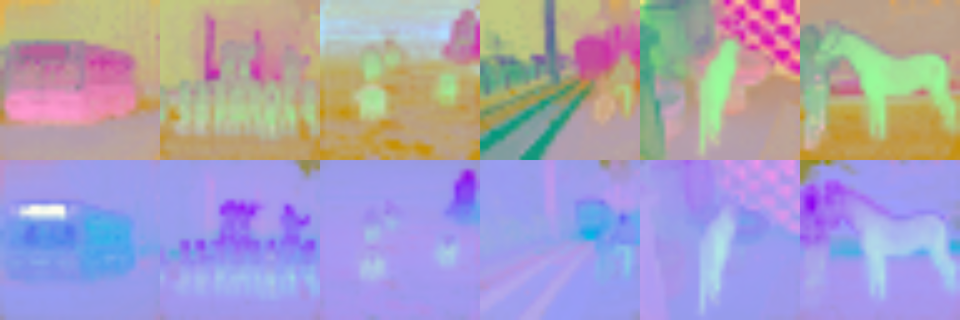}
      \end{center}
   \end{minipage}%

   \begin{minipage}[t]{0.03\linewidth}%
      \vspace{-68pt}
      \begin{sideways}\scriptsize{\textbf{\textsf{Eigs (V-V Graph)}}}\end{sideways}
   \end{minipage}%
   \begin{minipage}[t]{0.95\linewidth}%
      \begin{center}
         \includegraphics[width=1.0\linewidth]{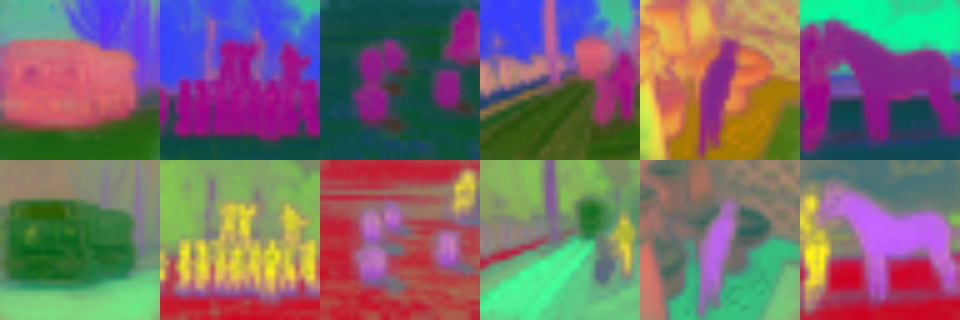}
      \end{center}
   \end{minipage}%
   \end{center}
   \vspace{-1.0em}
   \caption{%
      \textbf{Extracted eigenvectors on COCO for both graph choices.}
      We visualize selected components of $\mX_\text{ortho}$, sorted by decreasing eigenvalue.  Three eigenvectors at
      a time are rendered as RGB images.  In the Q-K case, the first set of eigenvectors describes general scene
      spatial layout in terms of ground, subject, background, and sky.  The second finds top-to-bottom part separation
      within objects.  In the V-V case, the first set of eigenvectors partitions the image into coarse semantics like
      trees, ground, and sky, while the second set recognizes finer-grained categories and groups individual
      objects like people, animals, and vehicles.%
   }%
   \label{fig:coco_viz}
\end{figure}

In Figure~\ref{fig:model_comparison}, we show that features and regions extracted from different models are quite
structured, aligning well with object boundaries.  We quantify region quality by measuring their oracle overlap with
semantic segmentation labels.  This gives a sense as to how well attention layers inside models decompose images along
semantic axes.  We perform this analysis on PASCAL VOC~\citep{everingham2015pascal}, which has 20 foreground classes
and 1449 validation images.  We score results with the metric of mean intersection over union (mIoU) between
regions and labels.  Each region is assigned to the ground-truth label it overlaps with the most.

\vspace{0.25em}
\noindent
\textbf{Results and Analysis.}
Table~\ref{tab:oracle_decoding} presents results demonstrating that our approach consistently outperforms all methods
to which we compare, across various backbone models.  For DINO, we show that directly clustering the final layer
features using K-Means yields decent performance.  This is likely due to the discriminative nature of DINO's final
representation, which makes a straightforward decoding strategy sufficient for generating satisfactory regions.
However, direct clustering fares much worse on other models with different training objectives.

Additionally, we observe that Normalized Cut~\citep{shi2000normalized} (Ncut) is highly sensitive to the underlying
graph, and its performance deteriorates significantly when switching from the graph of final features to the final
attention matrix.  A related approach, MaskCut~\citep{wang2023cut}, solves Ncut on a binarized graph to extract
foreground objects.  However, this operation results in the loss of finer-grained information, which is crucial for
segmentation tasks.  In contrast, our method is less sensitive to the quality of a single graph because we
simultaneously perform spectral clustering over a set of affinity matrices.  When comparing our method on models that
are not trained to produce discriminative features as their final output, such as MAE and CLIP, we observe an even
more substantial improvement.

\begin{figure}[t]
   \begin{center}
   \begin{minipage}[t]{0.03\linewidth}%
      \vspace{-30pt}
      \begin{sideways}\scriptsize{\textbf{\textsf{Image}}}\end{sideways}
   \end{minipage}%
   \begin{minipage}[t]{0.95\linewidth}%
      \begin{center}
         \includegraphics[width=1.0\linewidth]{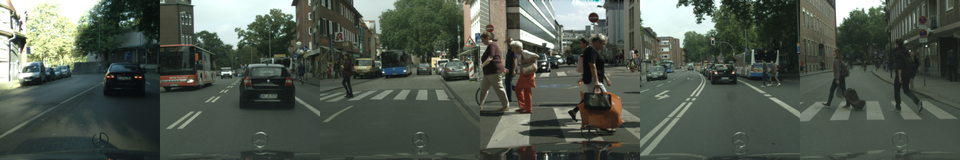}
      \end{center}
   \end{minipage}%

   \begin{minipage}[t]{0.03\linewidth}%
      \vspace{-68pt}
   \begin{sideways}\scriptsize{\textbf{\textsf{Eigs (Q-K Graph)}}}\end{sideways}
   \end{minipage}%
   \begin{minipage}[t]{0.95\linewidth}%
      \begin{center}
         \includegraphics[width=1.0\linewidth]{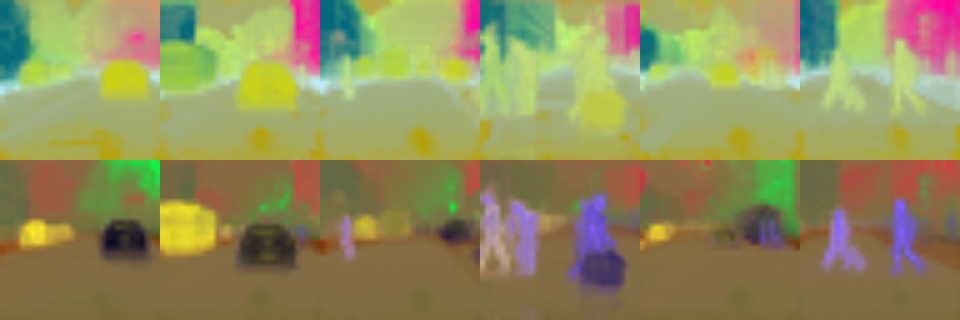}
      \end{center}
   \end{minipage}%

   \begin{minipage}[t]{0.03\linewidth}%
      \vspace{-68pt}
      \begin{sideways}\scriptsize{\textbf{\textsf{Eigs (V-V Graph)}}}\end{sideways}
   \end{minipage}%
   \begin{minipage}[t]{0.95\linewidth}%
      \begin{center}
         \includegraphics[width=1.0\linewidth]{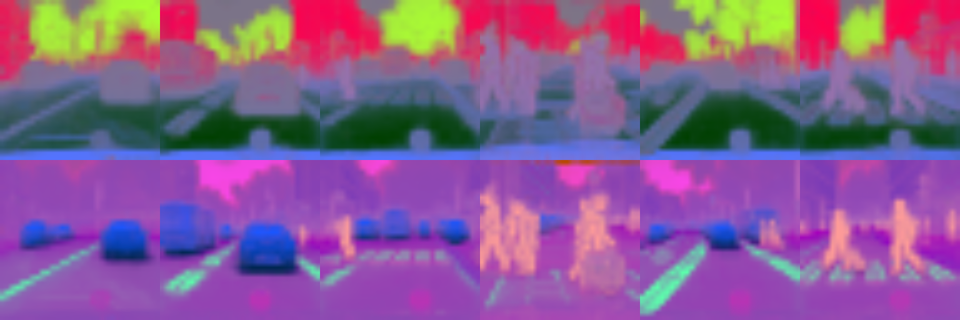}
      \end{center}
   \end{minipage}%
   \end{center}
   \vspace{-1.0em}
   \caption{%
      \textbf{Extracted eigenvectors on Cityscapes for both graph choices.}
      We visualize selected components of $\mX_\text{ortho}$, sorted by decreasing eigenvalue.  Three eigenvectors at
      a time are rendered as RGB images.  In the Q-K case, eigenvectors detect the scene spatial layout and indicate
      how far left or right buildings, cars, trees, and people are.  In the V-V case, eigenvectors perform semantic
      recognition and separate trees and buildings from road, and distinguish cars, people, and road markings.%
   }%
   \label{fig:cityscapes_viz}
\end{figure}

In Table \ref{tab:ablate_layer_time}, we provide ablation studies on the choices of layer for feature extraction and maximum noise level.  Stable diffusion has 16 attention layers 
with resolutions from 64x64 to 8x8. Our default is All (1-16), $t_\text{max}$ = 500. We conduct experiments on a per-image region extraction setting with 200 images from the PASCAL VOC validation set. Although our main experiment utilizes features from all layers, making minimal assumptions about layer-wise feature distribution, we find that using only decoder layers and a middle noise level yields better results.

To further evaluate the region quality irrespective of decoding choices, Figure~\ref{fig:main_miou_vs_k} shows
the mIoU with varying choice of $K$.  We see that the high quality of regions persists across choices, even when
compared with baselines.

Our per-image regions can find broad applicability in a variety of segmentation tasks.  For first-step
proof-of-concepts, see Appendices~\ref{app:clip_open_vocab_seg} and~\ref{app:instance_seg}.

\subsection{Full-Dataset Region Extraction}
\label{sec:full_dataset_results}

Our method can effectively extract regions within images.  Can it examine relationships across images?  To probe
different kinds of encoded information, we take the best model of the previous section, Stable Diffusion, as a
case-study.  We compare the query/key (Q-K) dataset-level graph with the value/value (V-V) dataset-level graph, as
described in Section~\ref{sec:dataset_extension}.  Results show a surprisingly structured split, where Q-K encodes
spatial information and V-V encodes semantic information, which we can use for tasks like unsupervised semantic
segmentation.

\vspace{0.25em}
\noindent
\textbf{Experimental Setup.}
For constructing graphs, we follow the method in Section~\ref{sec:dataset_extension}.  For efficiency, we concatenate
features at each head into a single vector instead of considering heads independently.  We select one attention block
in the middle block and the first 6 attention blocks in the up-sampling blocks, resulting in a total of 7 attention
matrices.  We choose channel number $C = 50$, cross-image connections $c_\text{inter} = c_\text{intra} = 10$, noise
level $t \in \mathcal{U}[20, 300)$, and optimize using Adam~\citep{kingma2014adam} with a learning rate of 1e-2, and a
batch size of 160 images over 4 GPUs for 2100 iterations.  When clustering, we choose $K$ to be the number of labels of
the relevant task.  More details are in Appendix~\ref{app:full_dataset_details}.

\subsubsection{Qualitative Analysis}
We show qualitative results for both Q-K and V-V graphs on COCO~\citep{lin2014microsoft} in Figure~\ref{fig:coco_viz}
and Cityscapes~\citep{cordts2016cityscapes} in Figure~\ref{fig:cityscapes_viz}.

Across datasets, we observe that the Q-K graph appears to encode spatial relationships.  On Cityscapes, which has a
clear spatial layout at the scene level, the learned eigenvectors effectively separate buildings, cars, people, and
trees into left/right subgroups.  For the more complex dataset COCO, which lacks fixed spatial patterns at the scene
level, the eigenvectors uncover spatial correlations first in terms of ground, subject, and background, and then
part-like correlations within objects from top-to-bottom.

By contrast, features from the V-V graph group objects semantically.  In COCO, we observe that eigenvectors encode
semantic structure hierarchically: the first set of eigenvectors focuses on distinguishing scene-level semantics
(\eg~ground, sky, trees) while overlooking differentiating foreground objects.  The next set of eigenvectors groups
foreground objects like people, animals, and vehicles.  In Cityscapes, the initial
set captures broad scene-level semantics, including trees, houses, and the egocentric vehicle, and can differentiate
between road and sidewalk.  The following set groups cars, people, and road markings.  More examples are available in
Appendix~\ref{app:qualitative}.

The qualitative differences between eigenvectors stemming from the Q-K and V-V graphs suggest a clear split in the way
the model processes information from images.  In attention layers, queries and keys are used to form patch-wise
relationships, which then modulate the values propagated to the next stage.  It appears that the model learns to split
representation into spatial and semantic branches as a convenient solution for taking advantage of this computational
structure.  This is perhaps surprising, as the projection from features to queries or keys or values need not split
information in such a clean fashion.  The discrepancy in the behavior between features from these different graphs
motivates us to further evaluate their performance in separate semantic and spatial benchmarks.

\subsubsection{Quantitative Analysis for `What' Pathway}
\label{sec:what_pathway}

\begin{table}
   \begin{center}
   \footnotesize
   \setlength{\tabcolsep}{11.5pt}
   \begin{tabular}{@{}l|c|c@{}}
      \bf \multirow{2}{*}{Method} & \multicolumn{2}{c}{\bf Results (mIOU)}\\
      & \bf COCO-Stuff-27 & \bf Cityscapes \\
      \HLINE
      MoCo v2 \citep{chen2020improved}   & 4.4 & - \\
      IIC \citep{ji2019invariant}  & 6.7 & 6.1\\
      DSM \citep{Melas-Kyriazi_2022_CVPR} (ViT-B/8) & 8.9 & - \\
      Modified DC \citep{cho2021picie} & 9.8 & 7.4 \\
      PiCIE \citep{cho2021picie}  &13.8 & 12.3 \\
      PiCIE+H \citep{cho2021picie}  & 14.4 & - \\
      ACSeg \citep{li2022adapting} & 16.4 & - \\
      HP \citep{seong2023leveraging} (ViT-S/8) & 24.6 & 18.4 \\
      STEGO \citep{hamilton2022unsupervised} (ViT-B/8) & 26.8 & 18.2 \\
      STEGO \citep{hamilton2022unsupervised}+CRF (ViT-B/8) & \bf 28.2 & \bf 21.0\\
      \hline
      Ours (V-V graph) &  25.4 & 16.2 \\
      Ours (V-V graph)+CRF &  27.1 & 16.9 \\
   \end{tabular}
   \end{center}
   \vspace{-1.0em}
   \caption{%
      \textbf{Unsupervised semantic segmentation results on COCO-Stuff-27 and Cityscapes.}
      We observe that the V-V graph features outperform those of prior works and achieve competitive performance
      compared to the strong STEGO method, which utilizes discriminative DINO~\cite{caron2021emerging} features and a
      complex two-stage global nearest-neighbor strategy.  Conversely, our method employs representations from a
      generative model and collects neighbors solely from the minibatch, a simpler and more scalable approach.%
   }%
   \label{tab:kmeans_coco}
\end{table}

To quantify semantic segmentation ability in the V-V graph, we evaluate our extracted segmentation on two common
unsupervised semantic segmentation tasks: COCO-Stuff~\citep{lin2014microsoft, caesar2018coco} and
Cityscapes~\citep{cordts2016cityscapes}.  We follow the preprocessing protocol as adopted in PiCIE~\citep{cho2021picie}
and STEGO~\citep{hamilton2022unsupervised}.  We optimize $\mX$ on the validation set, where images are first resized
so the minor edge is 320px and then cropped in the center to produce square images.  We choose $K = 27$, the number of
ground-truth categories in both datasets, for K-Means over $\mX$, and then use greedy matching to align the cluster
assignments with the ground truth.  We report results with mIoU and compare to other methods in
Table~\ref{tab:kmeans_coco}, and examine feature choice and decoding protocol in Table~\ref{tab:kmeans_coco_decoding}.
More details are in Appendix~\ref{app:full_dataset_details}.

In Table~\ref{tab:kmeans_coco}, our method significantly outperforms many other methods and is comparable to
STEGO~\citep{hamilton2022unsupervised}.  This is quite surprising, as STEGO~\citep{hamilton2022unsupervised} adopts a
sophisticated two-stage dataset-wise nearest-neighbor searching procedure, while our method only considers connections
within the mini-batch, a strategy with noisy signal but with better scalability.  STEGO~\citep{hamilton2022unsupervised}
also benefits from the discriminative representations of DINO~\citep{caron2021emerging}, while our backbone, Stable
Diffusion~\citep{rombach2022high}, is generative.

Table~\ref{tab:kmeans_coco_decoding} reports results on directly clustering the most semantic representations of Stable
Diffusion~\citep{rombach2022high}, which are the features of the 2nd upsampling block with timestep
$t=250$~\citep{tang2023emergent}.  In this comparison, DINO~\citep{caron2021emerging} features are better than Stable
Diffusion, likely due to their discriminative properties, but our method greatly narrows the gap.

\begin{table}
   \footnotesize
   \setlength{\tabcolsep}{5.25pt}
   \begin{tabular}{@{}l|c|c|c|c@{}}
      \bf \multirow{2}{*}{Method} & \multicolumn{2}{c|}{\bf COCO-Stuff-27} & \multicolumn{2}{c}{\bf Cityscapes}\\
      & Greedy & Hungarian & Greedy & Hungarian\\
      \HLINE
      K-Means (SD \citep{rombach2022high}) & 9.2 & 8.6 & 12.4 &8.1 \\
      K-Means (DINO \citep{caron2021emerging}) & 13.7 & 13.0 & 13.3& 8.7\\
      K-Means (STEGO \citep{hamilton2022unsupervised}) &  26.6 & 24.0& 15.8 & 14.9\\
      STEGO \citep{hamilton2022unsupervised} & \bf 27.0 & \bf 26.5 & \bf 16.6 & \bf 18.2\\
      \hline
      Ours (Q-K graph) &  12.4 & 10.9 & 10.91 & 9.7 \\
      Ours (V-V graph) &  25.4 & 23.2 & 16.2 &  11.4 \\
   \end{tabular}
   \caption{%
      \textbf{Ablations for unsupervised semantic segmentation.}
      We test multiple sources of features for clustering on the validation set only, and vary the decoding pipeline
      for evaluation.  With greedy decoding, our features are comparable, but with Hungarian matching STEGO is
      stronger.  We also see that the Q-K graph encodes far less semantic information than the V-V graph, supporting a
      semantic/spatial decomposition.  The discrepancy between K-Means (STEGO) and STEGO numbers here is due to
      restricting clustering to the validation set.%
   }%
   \label{tab:kmeans_coco_decoding}
\end{table}

\subsubsection{Quantitative Analysis for `Where' Pathway}\label{sec:where_pathway}

Here, we design two experiments to quantify the positional information in the Q-K graph for the `where' pathway.  Our
first experiment aims to measure the amount of positional information contained within the features.  In this setting,
we train a linear head on top of the features and attempt to regress the corresponding grid position of each
pixel/patch.  We call this task ``coordinate regression".

\begin{figure}[t]
   \begin{center}
      \scriptsize
      \begin{subfigure}[t]{0.325\linewidth}
         \centering
         \includegraphics[width=\linewidth]{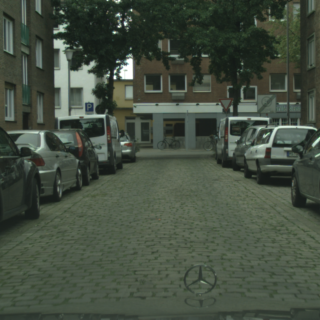}
         \caption{Image}
      \end{subfigure}
      \hfill
      \begin{subfigure}[t]{0.325\linewidth}
         \centering
         \includegraphics[width=\linewidth]{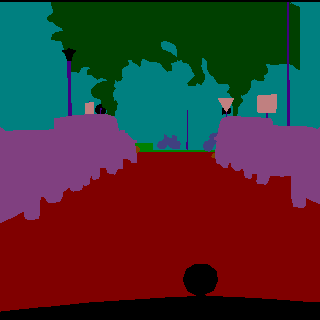}
         \caption{Semantic Label}
      \end{subfigure}
      \hfill
      \begin{subfigure}[t]{0.325\linewidth}
         \centering
         \includegraphics[width=\linewidth]{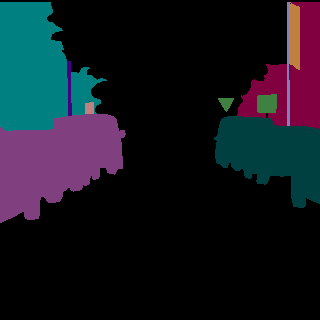}
         \caption{Processed Label}
      \end{subfigure}%
   \end{center}
   \vspace{-1.0em}
   \caption{%
      \textbf{Spatial semantic segmentation task.}
      We generate labels for ``spatial semantic segmentation'' by splitting the semantic labels into left/right
      subgroups, followed by filtering out small regions and ambiguous regions close to the image center.%
   }%
   \label{fig:cityscape_spatial_label}
\end{figure}

The second experiment aims to evaluate whether this spatial information is present at the semantic level.  In this
case, we benchmark on an unsupervised semantic segmentation task by further partitioning the semantic annotations into
left and right subgroups.  For this purpose, we process the ground-truth annotations by first identifying disconnected
regions for each category of segmentation annotation and then scoring each region based on its pixel distance to the
image's left/right border.  We filter out smaller regions with pixel counts less than 50 and ambiguous regions located
close to the image center.  We call this task ``spatial semantic segmentation" and showcase the original and processed
semantic segmentation maps in Figure~\ref{fig:cityscape_spatial_label}.  We follow the exact same evaluation protocol
as used in the experiment for the `what' pathway.

The results of both experiments are presented in Table~\ref{tab:spatial_exp}, where we compare with STEGO features,
DINO final-layer features, and ground-truth semantic segmentation labels.  In both experiments, our approach with the
Q-K graph outperforms both STEGO and the variants with the V-V graph.

Results on coordinate regression suggest that $\mX$ from the Q-K graph contains rich spatial information for processing
the `where' pathway.  However, both STEGO and the V-V graph group pixels only by semantic similarity and remove spatial
information from the final representation.  DINO performs well on regression likely due to position embeddings.

We further verify the spatial information content of the Q-K graph by examining the results of spatial semantic
segmentation.  We see that these features are strongest in this task.  Compared to results on unsupervised semantic
segmentation (Table~\ref{tab:kmeans_coco}), the strong performance of features from the V-V graph and STEGO
deteriorates due to failure to reason about spatial structure.  DINO features also fail in this task, likely as spatial
information is not as strong a signal as semantics for discriminating between images.  These results, along with those
in Section~\ref{sec:what_pathway}, show that our approach can scale efficiently to extract both spatial and semantic
relationships across images.

\begin{table}
   \footnotesize
   \setlength{\tabcolsep}{8.5pt}
   \begin{tabular}{@{}l|c|c@{}}
      \bf \shortstack{Method\\~} & \shortstack{Coordinate\\Regression (MSE) $\downarrow$} &
                                   \shortstack{Spatial Semantic\\Segmentation (mIOU) $\uparrow$}\\
      \HLINE
      DINO & \bf 3.2 & 6.0 \\
      GT semantic label & 72.0 & - \\
      \hline
      STEGO & 42.4 & 6.9\\
      Ours (V-V graph) & 43.1 & 5.1 \\
      Ours (Q-K graph)& \bf 19.5 & \bf 10.1\\
   \end{tabular}
   \caption{\textbf{Results for evaluating `where' pathway on spatial structures}. Ours (Q-K graph) outperforms STEGO and ours (V-V graph) in both benchmarks suggesting the Q-K graph contains richer spatial information at object levels. Though DINO can trivially recover the spatial coordinates through positional embeddings, it fails to leverage that information for segmentation.}
   \label{tab:spatial_exp}
\end{table}

\section{Discussion}

We present an approach for extracting information from a neural network's activations.  Unlike prior work, our method
examines the whole of a network, without needing to guess which part of the model contains relevant features.  Our
approach resembles classic spectral clustering, but gains scalability to dataset-level analysis by approximating a
solution using gradient-based optimization.

Deployed as a mechanism for extracting image segmentation from large pre-trained models, we observe robust
performance in producing regions from a wide variety of source models, including high quality semantic segmentations
obtained from a Stable Diffusion model.  Deployed as an analysis tool, we gain new insight into how vision models
with attention layers utilize key, query, and value vectors to coordinate the flow of spatial and semantic
information, and disentangle `what' and `where' pathways within these deep networks.

Our approach could be the first example in a new class of optimization-centric techniques for peering into
the inner workings of deep networks.  Future research could repurpose other computationally intensive, but
scalable, classic machine learning tools to the analysis of network behavior.

\vspace{0.5em}
\noindent
{\small{\textbf{Acknowledgments.} NSF GRFP (1754881) supported D.~Yunis.}}


{
   \small
   \bibliographystyle{ieeenat_fullname}
   \bibliography{main}
}

\appendix
\newpage

\begin{figure*}[t!]
      \begin{center}
            \begin{minipage}[t]{0.03\linewidth}%
                \vspace{-15pt}
                  \begin{sideways}\scriptsize{\textbf{\textsf{Img}}}\end{sideways}
            \end{minipage}%
            \begin{minipage}[t]{0.95\linewidth}%
                  \begin{center}
                     \includegraphics[width=1.0\linewidth]{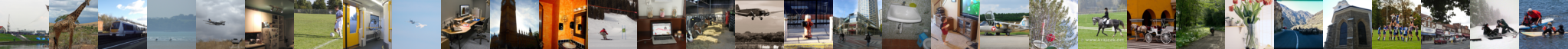}
                     \\
                     \vspace{-10pt}
                     \textcolor{gray}{\rule{\linewidth}{0.75pt}}
                  \end{center}
            \end{minipage}%
            
            \begin{minipage}[t]{0.03\linewidth}%
                \vspace{-58pt}
                  \begin{sideways}\scriptsize{\textbf{\textsf{Eigs (Q-K Graph)}}}\end{sideways}
            \end{minipage}%
            \begin{minipage}[t]{0.95\linewidth}%
                  \begin{center}
                     \includegraphics[width=1.0\linewidth]{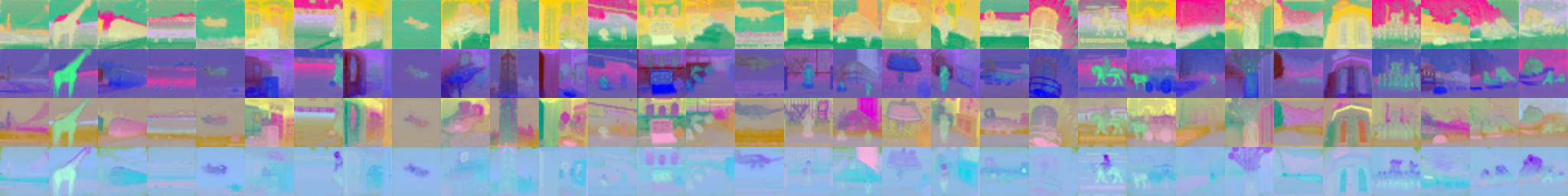}
                     \\
                     \vspace{-10pt}
                     \textcolor{gray}{\rule{\linewidth}{0.75pt}}
                  \end{center}
            \end{minipage}%
            
            \begin{minipage}[t]{0.03\linewidth}%
                \vspace{-58pt}
                  \begin{sideways}\scriptsize{\textbf{\textsf{Eigs (V-V Graph)}}}\end{sideways}
            \end{minipage}%
            \begin{minipage}[t]{0.95\linewidth}%
                  \begin{center}
                     \includegraphics[width=1.0\linewidth]{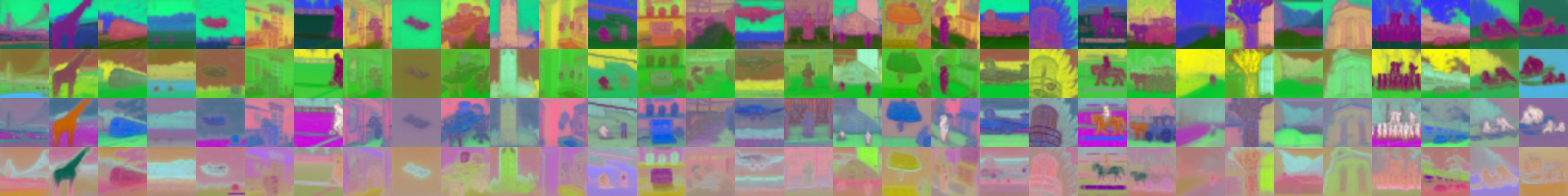}
                  \end{center}
            \end{minipage}%
    \vspace{-14pt}
           \begin{center}
                     \textcolor{gray}{\rule{\linewidth}{1.5pt}}
                  \end{center}
    \vspace{-3pt}
            \begin{minipage}[t]{0.03\linewidth}%
                \vspace{-15pt}
                  \begin{sideways}\scriptsize{\textbf{\textsf{Img}}}\end{sideways}
            \end{minipage}%
            \begin{minipage}[t]{0.95\linewidth}%
                  \begin{center}
                     \includegraphics[width=1.0\linewidth]{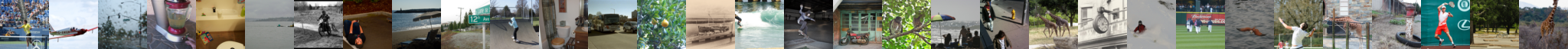}
                     \\
                     \vspace{-10pt}
                     \textcolor{gray}{\rule{\linewidth}{0.75pt}}
                  \end{center}
            \end{minipage}%
            
            \begin{minipage}[t]{0.03\linewidth}%
                \vspace{-58pt}
                  \begin{sideways}\scriptsize{\textbf{\textsf{Eigs (Q-K Graph)}}}\end{sideways}
            \end{minipage}%
            \begin{minipage}[t]{0.95\linewidth}%
                  \begin{center}
                     \includegraphics[width=1.0\linewidth]{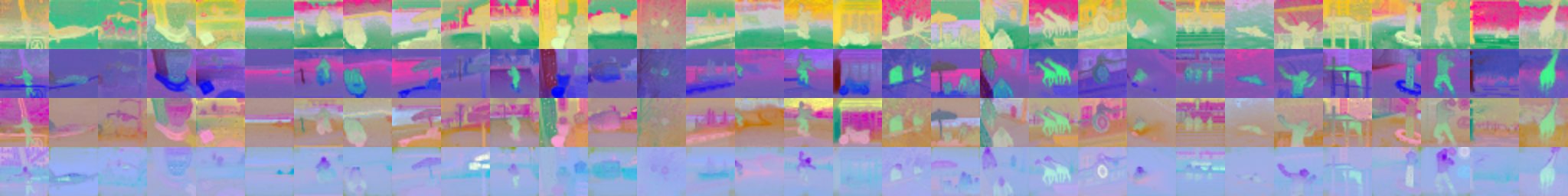}
                     \\
                     \vspace{-10pt}
                     \textcolor{gray}{\rule{\linewidth}{0.75pt}}
                  \end{center}
            \end{minipage}%
            
            \begin{minipage}[t]{0.03\linewidth}%
                \vspace{-58pt}
                  \begin{sideways}\scriptsize{\textbf{\textsf{Eigs (V-V Graph)}}}\end{sideways}
            \end{minipage}%
            \begin{minipage}[t]{0.95\linewidth}%
                  \begin{center}
                     \includegraphics[width=1.0\linewidth]{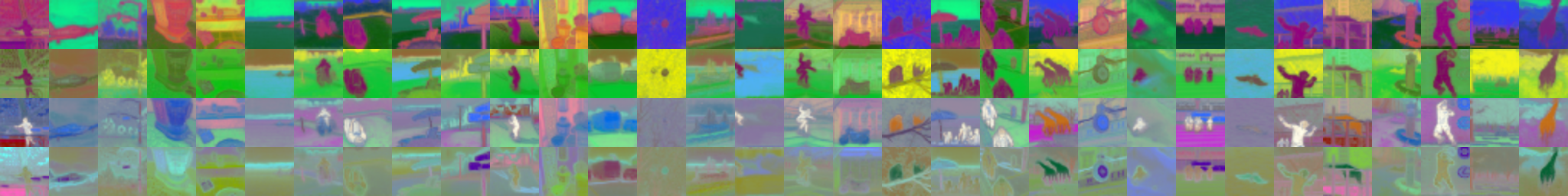}
                  \end{center}
            \end{minipage}%
    \vspace{-14pt}
           \begin{center}
                     \textcolor{gray}{\rule{\linewidth}{1.5pt}}
                  \end{center}
    \vspace{-3pt}
            \begin{minipage}[t]{0.03\linewidth}%
                \vspace{-15pt}
                  \begin{sideways}\scriptsize{\textbf{\textsf{Img}}}\end{sideways}
            \end{minipage}%
            \begin{minipage}[t]{0.95\linewidth}%
                  \begin{center}
                     \includegraphics[width=1.0\linewidth]{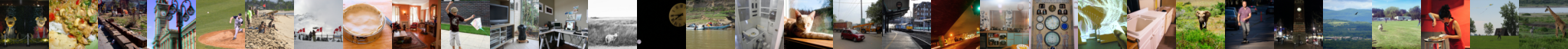}
                     \\
                     \vspace{-10pt}
                     \textcolor{gray}{\rule{\linewidth}{0.75pt}}
                  \end{center}
            \end{minipage}%
            
            \begin{minipage}[t]{0.03\linewidth}%
                \vspace{-58pt}
                  \begin{sideways}\scriptsize{\textbf{\textsf{Eigs (Q-K Graph)}}}\end{sideways}
            \end{minipage}%
            \begin{minipage}[t]{0.95\linewidth}%
                  \begin{center}
                     \includegraphics[width=1.0\linewidth]{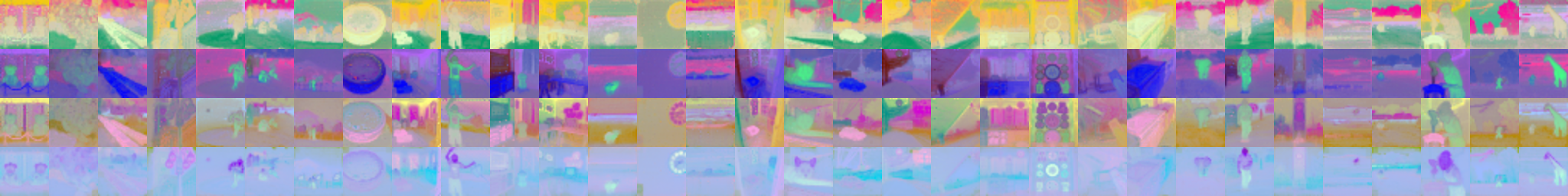}
                     \\
                     \vspace{-10pt}
                     \textcolor{gray}{\rule{\linewidth}{0.75pt}}
                  \end{center}
            \end{minipage}%
            
            \begin{minipage}[t]{0.03\linewidth}%
                \vspace{-58pt}
                  \begin{sideways}\scriptsize{\textbf{\textsf{Eigs (V-V Graph)}}}\end{sideways}
            \end{minipage}%
            \begin{minipage}[t]{0.95\linewidth}%
                  \begin{center}
                     \includegraphics[width=1.0\linewidth]{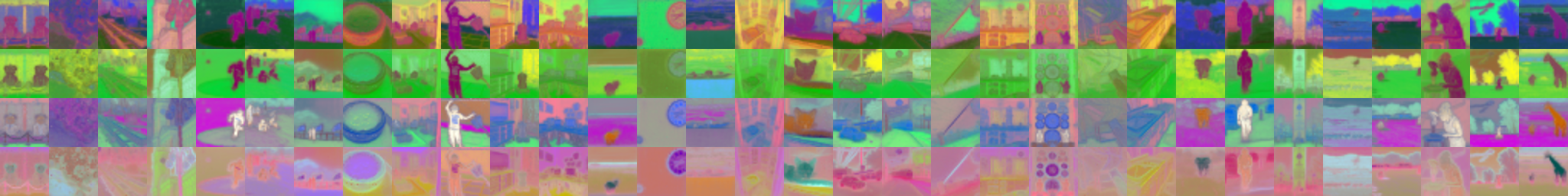}
                  \end{center}
            \end{minipage}%
    \vspace{-14pt}
           \begin{center}
                     \textcolor{gray}{\rule{\linewidth}{1.5pt}}
                  \end{center}
    \vspace{-3pt}
            \begin{minipage}[t]{0.03\linewidth}%
                \vspace{-15pt}
                  \begin{sideways}\scriptsize{\textbf{\textsf{Img}}}\end{sideways}
            \end{minipage}%
            \begin{minipage}[t]{0.95\linewidth}%
                  \begin{center}
                     \includegraphics[width=1.0\linewidth]{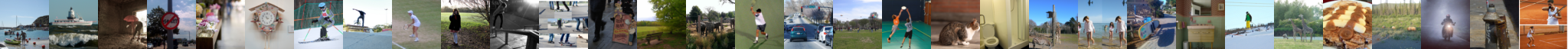}
                     \\
                     \vspace{-10pt}
                     \textcolor{gray}{\rule{\linewidth}{0.75pt}}
                  \end{center}
            \end{minipage}%
            
            \begin{minipage}[t]{0.03\linewidth}%
                \vspace{-58pt}
                  \begin{sideways}\scriptsize{\textbf{\textsf{Eigs (Q-K Graph)}}}\end{sideways}
            \end{minipage}%
            \begin{minipage}[t]{0.95\linewidth}%
                  \begin{center}
                     \includegraphics[width=1.0\linewidth]{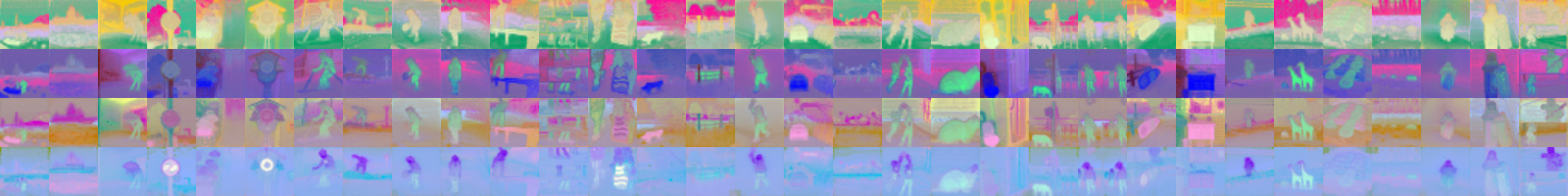}
                     \\
                     \vspace{-10pt}
                     \textcolor{gray}{\rule{\linewidth}{0.75pt}}
                  \end{center}
            \end{minipage}%
            
            \begin{minipage}[t]{0.03\linewidth}%
                \vspace{-58pt}
                  \begin{sideways}\scriptsize{\textbf{\textsf{Eigs (V-V Graph)}}}\end{sideways}
            \end{minipage}%
            \begin{minipage}[t]{0.95\linewidth}%
                  \begin{center}
                     \includegraphics[width=1.0\linewidth]{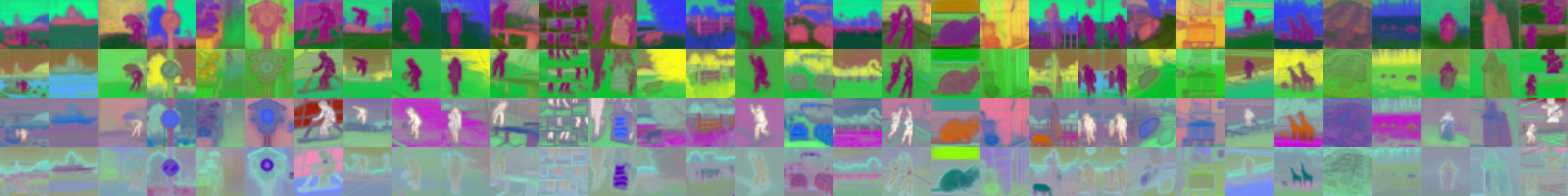}
                  \end{center}
            \end{minipage}%
        \caption{\textbf{More examples of extracted eigenvectors on COCO for both graph choices.} We visualize selected components of $\mX_\text{ortho}$, sorted by decreasing eigenvalue. Three eigenvectors at a time are rendered as RGB images.}
        \label{fig:appendix_coco_viz}
      \end{center}
 \end{figure*}

\begin{figure*}[t!]
      \begin{center}
            \begin{minipage}[t]{0.03\linewidth}%
                \vspace{-15pt}
                  \begin{sideways}\scriptsize{\textbf{\textsf{Img}}}\end{sideways}
            \end{minipage}%
            \begin{minipage}[t]{0.95\linewidth}%
                  \begin{center}
                     \includegraphics[width=1.0\linewidth]{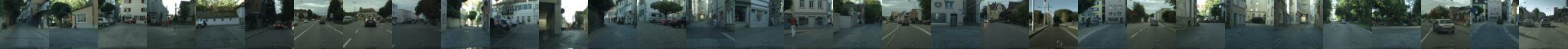}
                     \\
                     \vspace{-10pt}
                     \textcolor{gray}{\rule{\linewidth}{0.75pt}}
                  \end{center}
            \end{minipage}%
            
            \begin{minipage}[t]{0.03\linewidth}%
                \vspace{-58pt}
                  \begin{sideways}\scriptsize{\textbf{\textsf{Eigs (Q-K Graph)}}}\end{sideways}
            \end{minipage}%
            \begin{minipage}[t]{0.95\linewidth}%
                  \begin{center}
                     \includegraphics[width=1.0\linewidth]{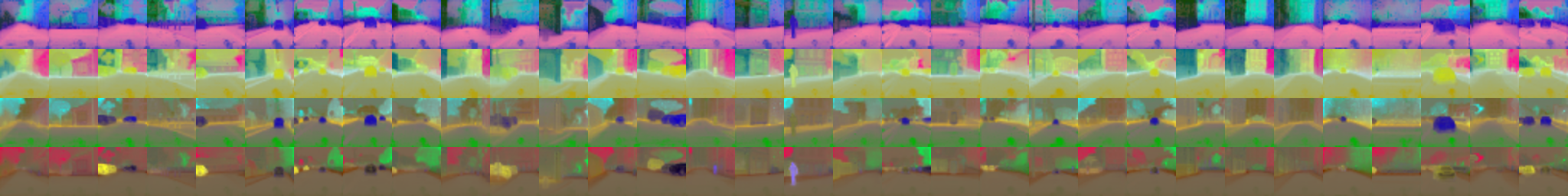}
                     \\
                     \vspace{-10pt}
                     \textcolor{gray}{\rule{\linewidth}{0.75pt}}
                  \end{center}
            \end{minipage}%
            
            \begin{minipage}[t]{0.03\linewidth}%
                \vspace{-58pt}
                  \begin{sideways}\scriptsize{\textbf{\textsf{Eigs (V-V Graph)}}}\end{sideways}
            \end{minipage}%
            \begin{minipage}[t]{0.95\linewidth}%
                  \begin{center}
                     \includegraphics[width=1.0\linewidth]{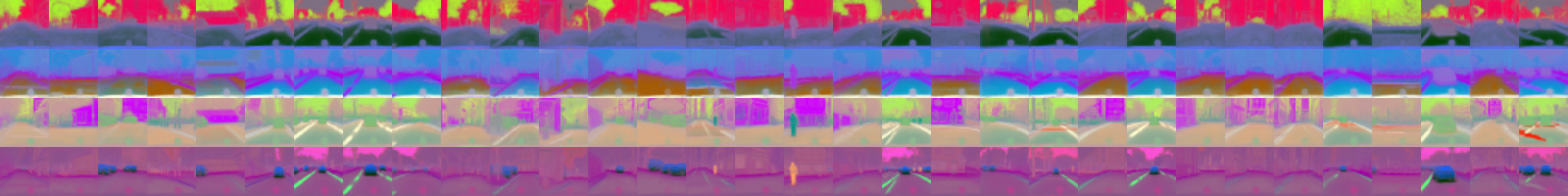}
                  \end{center}
            \end{minipage}%
    \vspace{-14pt}
           \begin{center}
                     \textcolor{gray}{\rule{\linewidth}{1.5pt}}
                  \end{center}
    \vspace{-3pt}
            \begin{minipage}[t]{0.03\linewidth}%
                \vspace{-15pt}
                  \begin{sideways}\scriptsize{\textbf{\textsf{Img}}}\end{sideways}
            \end{minipage}%
            \begin{minipage}[t]{0.95\linewidth}%
                  \begin{center}
                     \includegraphics[width=1.0\linewidth]{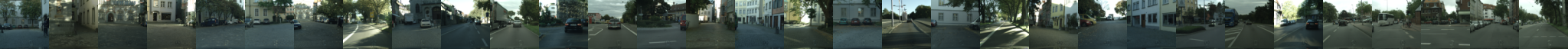}
                     \\
                     \vspace{-10pt}
                     \textcolor{gray}{\rule{\linewidth}{0.75pt}}
                  \end{center}
            \end{minipage}%
            
            \begin{minipage}[t]{0.03\linewidth}%
                \vspace{-58pt}
                  \begin{sideways}\scriptsize{\textbf{\textsf{Eigs (Q-K Graph)}}}\end{sideways}
            \end{minipage}%
            \begin{minipage}[t]{0.95\linewidth}%
                  \begin{center}
                     \includegraphics[width=1.0\linewidth]{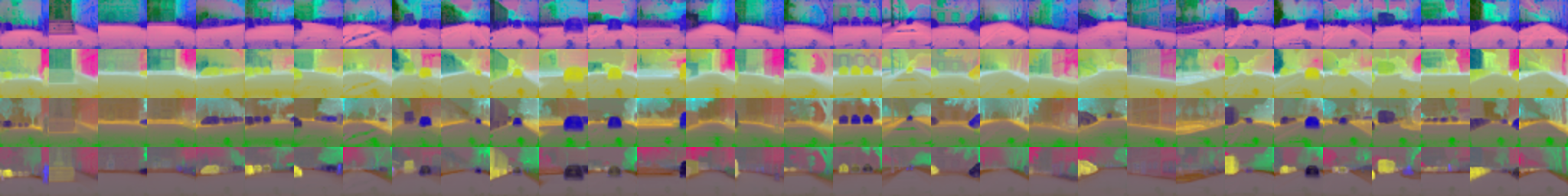}
                     \\
                     \vspace{-10pt}
                     \textcolor{gray}{\rule{\linewidth}{0.75pt}}
                  \end{center}
            \end{minipage}%
            
            \begin{minipage}[t]{0.03\linewidth}%
                \vspace{-58pt}
                  \begin{sideways}\scriptsize{\textbf{\textsf{Eigs (V-V Graph)}}}\end{sideways}
            \end{minipage}%
            \begin{minipage}[t]{0.95\linewidth}%
                  \begin{center}
                     \includegraphics[width=1.0\linewidth]{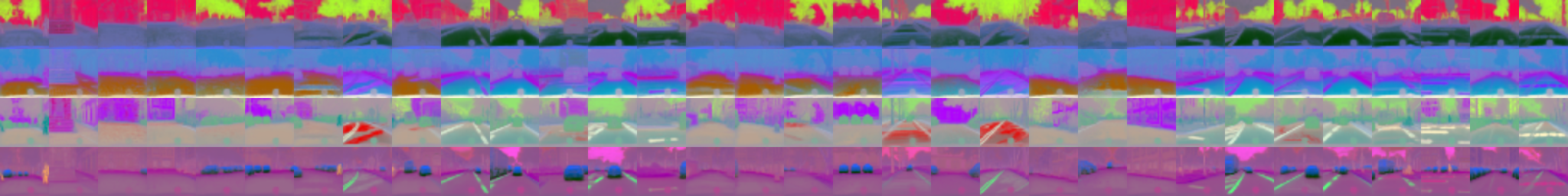}
                  \end{center}
            \end{minipage}%
    \vspace{-14pt}
           \begin{center}
                     \textcolor{gray}{\rule{\linewidth}{1.5pt}}
                  \end{center}
    \vspace{-3pt}
            \begin{minipage}[t]{0.03\linewidth}%
                \vspace{-15pt}
                  \begin{sideways}\scriptsize{\textbf{\textsf{Img}}}\end{sideways}
            \end{minipage}%
            \begin{minipage}[t]{0.95\linewidth}%
                  \begin{center}
                     \includegraphics[width=1.0\linewidth]{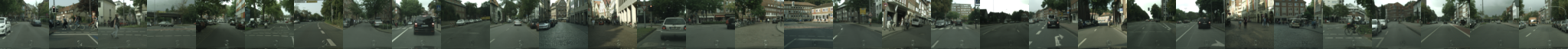}
                     \\
                     \vspace{-10pt}
                     \textcolor{gray}{\rule{\linewidth}{0.75pt}}
                  \end{center}
            \end{minipage}%
            
            \begin{minipage}[t]{0.03\linewidth}%
                \vspace{-58pt}
                  \begin{sideways}\scriptsize{\textbf{\textsf{Eigs (Q-K Graph)}}}\end{sideways}
            \end{minipage}%
            \begin{minipage}[t]{0.95\linewidth}%
                  \begin{center}
                     \includegraphics[width=1.0\linewidth]{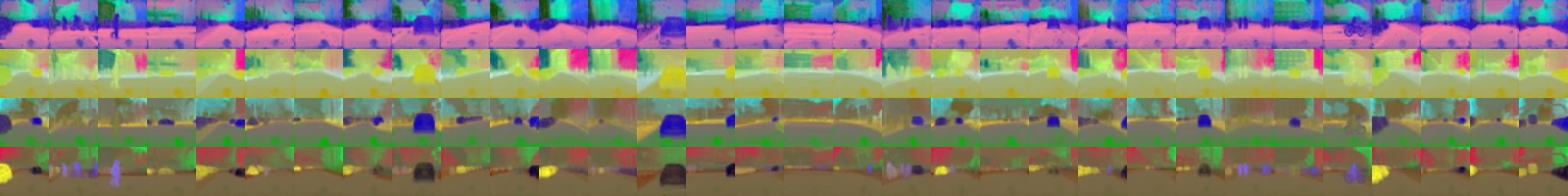}
                     \\
                     \vspace{-10pt}
                     \textcolor{gray}{\rule{\linewidth}{0.75pt}}
                  \end{center}
            \end{minipage}%
            
            \begin{minipage}[t]{0.03\linewidth}%
                \vspace{-58pt}
                  \begin{sideways}\scriptsize{\textbf{\textsf{Eigs (V-V Graph)}}}\end{sideways}
            \end{minipage}%
            \begin{minipage}[t]{0.95\linewidth}%
                  \begin{center}
                     \includegraphics[width=1.0\linewidth]{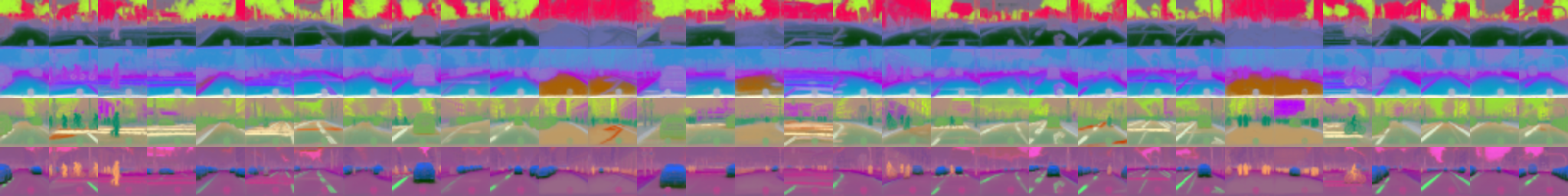}
                  \end{center}
            \end{minipage}%
    \vspace{-14pt}
           \begin{center}
                     \textcolor{gray}{\rule{\linewidth}{1.5pt}}
                  \end{center}
    \vspace{-3pt}
            \begin{minipage}[t]{0.03\linewidth}%
                \vspace{-15pt}
                  \begin{sideways}\scriptsize{\textbf{\textsf{Img}}}\end{sideways}
            \end{minipage}%
            \begin{minipage}[t]{0.95\linewidth}%
                  \begin{center}
                     \includegraphics[width=1.0\linewidth]{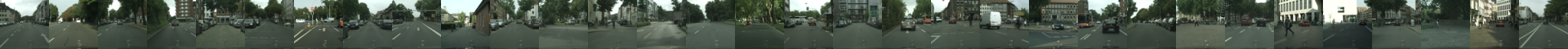}
                     \\
                     \vspace{-10pt}
                     \textcolor{gray}{\rule{\linewidth}{0.75pt}}
                  \end{center}
            \end{minipage}%
            
            \begin{minipage}[t]{0.03\linewidth}%
                \vspace{-58pt}
                  \begin{sideways}\scriptsize{\textbf{\textsf{Eigs (Q-K Graph)}}}\end{sideways}
            \end{minipage}%
            \begin{minipage}[t]{0.95\linewidth}%
                  \begin{center}
                     \includegraphics[width=1.0\linewidth]{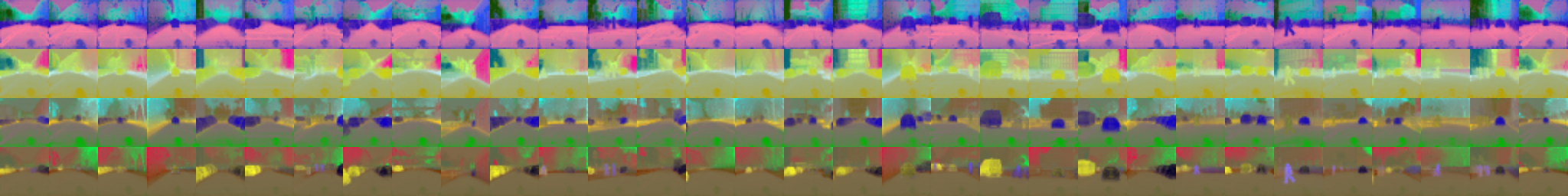}
                     \\
                     \vspace{-10pt}
                     \textcolor{gray}{\rule{\linewidth}{0.75pt}}
                  \end{center}
            \end{minipage}%
            
            \begin{minipage}[t]{0.03\linewidth}%
                \vspace{-58pt}
                  \begin{sideways}\scriptsize{\textbf{\textsf{Eigs (V-V Graph)}}}\end{sideways}
            \end{minipage}%
            \begin{minipage}[t]{0.95\linewidth}%
                  \begin{center}
                     \includegraphics[width=1.0\linewidth]{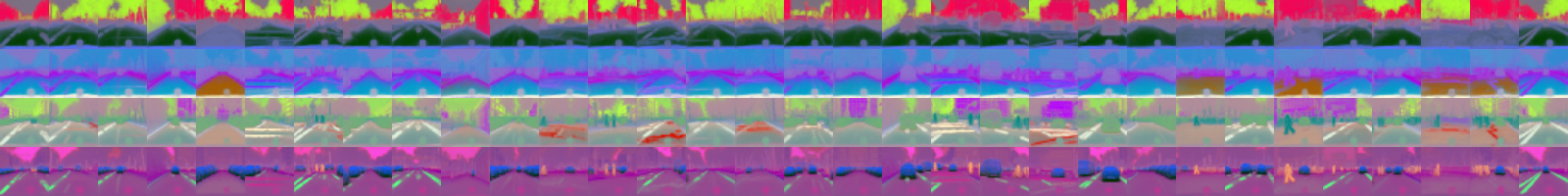}
                  \end{center}
            \end{minipage}%
        \caption{\textbf{More examples of extracted eigenvectors on Cityscapes for both graph choices.} We visualize selected components of $\mX_\text{ortho}$, sorted by decreasing eigenvalue. Three eigenvectors at a time are rendered as RGB images.}
        \label{fig:appendix_cityscape_viz}
      \end{center}
 \end{figure*}

\section{Additional Qualitative Results}\label{app:qualitative}

We provide additional visualizations of the extracted eigenvectors for both the COCO and Cityscapes datasets in
Figures~\ref{fig:appendix_coco_viz} and~\ref{fig:appendix_cityscape_viz}.  These visualizations follow the same
methodology as in Figures~\ref{fig:coco_viz} and~\ref{fig:cityscapes_viz}, where eigenvectors $\mU$ are rendered in
descending order in groups of three.  Each channel of the RGB image corresponds to the value of a particular
eigenvector at that coordinate.

\section{Per-Image Experimental Details}\label{app:per_image_details}

\begin{figure}[t]
   \begin{center}
      \includegraphics[width=0.9\linewidth]{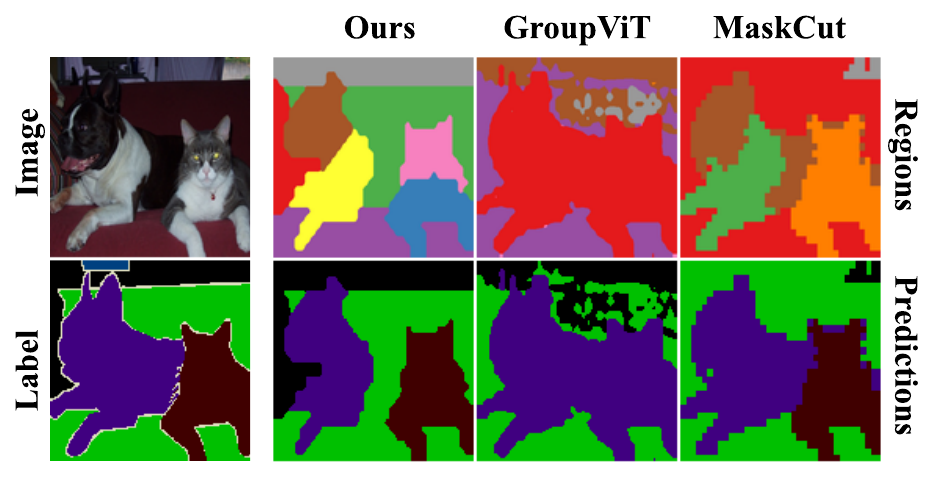}
   \end{center}
   \vspace{-1.0em}
   \caption{%
      \textbf{Examples of different segmentation methods on PASCAL VOC~\cite{everingham2015pascal}.}
      All methods besides GroupViT~\cite{xu2022groupvit} use DINO~\cite{caron2021emerging} features or attention.
      Ours can generate diversified regions while maintaining accurate object borders.  In contrast,
      GroupViT~\cite{caron2021emerging} tends to generate a noisy boundary, while MaskCut~\cite{wang2023cut} can miss
      subtle boundaries.%
   }%
   \label{fig:region_comparison}
\end{figure}

\subsection{Data Preprocessing}

For models besides Stable Diffusion~\cite{rombach2022high}, and Masked Autoencoder (MAE)~\cite{he2022masked} image
inputs are resized to have a short dimension of 448 pixels.  This requires change to the resolution of the learned
positional embeddings, which we do through bicubic interpolation, similar to MaskCLIP~\cite{zhou2022extract}.  In the
case of Stable Diffusion, we instead resize images to 512x512 to match the input dimensions of the original model.
For MAE, we resize images to 224x224, then upsample the internal query and key matrices to match the spatial resolution
of CLIP and DINO.

\subsection{Optimization}

We optimize features with Adam~\cite{kingma2014adam}, with a learning rate of 3e-4 or 1e-3, and default
PyTorch~\cite{paszke2019pytorch} betas $(0.9, 0.999)$.  We take a number of gradient steps to convergence that depends
on the model (1000 for CLIP, 2000 for others), but we typically find that 1000 steps is sufficient.  Timing information
for our method is available in Table~\ref{tab:per_image_timing}.

Unlike other models, where there is only a single set of attention matrices per image, the sampling of $t$ in the
forward pass of Stable Diffusion introduces more noise and significantly more computation into the optimization.  To
address this, we cache attention matrices in a buffer of 5 at a time, where the chance to sample a new set of attention
matrices is 1/4, and the oldest set in the buffer is replaced by this sample.  We also accumulate gradients for 20
backward passes before taking an optimizer step.

\subsection{Baselines}

To extract regions from TokenCut~\cite{wang2022tokencut} and MaskCut~\cite{wang2023cut}, a single affinity matrix is
required.  One choice is an affinity matrix constructed from features of the final layer, which is the original
proposed matrix for these methods.  Another is the final layer's attention matrix.  A third alternative is to compute
an average over all attention matrices across layers, so as to better compare to our method.  We found the third option
often led to an ill-conditioned matrix, which could not be solved.

Consequently, we present results for the first two choices.  For methods except TokenCut, we find best results with
$m=15$ eigenvectors.  For TokenCut we found the performance with $m=15$ to be subpar, so we use $m=8$ instead.
Quantitative results are available in Table~\ref{tab:oracle_decoding}.  Qualitative results comparing decoding methods
can be seen in Figure~\ref{fig:region_comparison}.  See Figure~\ref{fig:model_comparison} for comparison between
regions extracted from different models.

\subsection{Computational Cost}

\begin{table}
   \begin{center}
   \footnotesize
   \setlength{\tabcolsep}{20pt}
   \begin{tabular}{@{}l|c@{}}
      \bf Model & \bf Runtime (seconds) \\
      \HLINE
      Stable Diffusion 1.4 (w/ buffer)~\cite{rombach2022high} & 67 \\
      Stable Diffusion 1.4 (w/o buffer)~\cite{rombach2022high} & 155 \\
      DINO ViT-S/16~\cite{caron2021emerging} & 40\\
      MAE/CLIP ViT-B/16~\cite{radford2021learning} & 54
   \end{tabular}
   \end{center}
   \vspace{-1.0em}
   \caption{%
      \textbf{Computation time across models.}
      We benchmark region computation time for 1000 optimization steps using different models on an NVIDIA A40.
      1000 steps are often not required for good results, thus it may be possible to significantly accelerate the
      pipeline.%
   }%
   \label{tab:per_image_timing}
\end{table}

Table~\ref{tab:per_image_timing} shows the computational cost of running our method, benchmarked on an NVIDIA A40.
The extremely long computation time for Stable Diffusion is due to many evaluations of the model during optimization,
instead of simply caching the attention matrices from a single forward pass.

\section{Full-Dataset Experimental Details}\label{app:full_dataset_details}

\subsection{Data Preprocessing}

All experiments take place on COCO-Stuff~\citep{lin2014microsoft, caesar2018coco} and
Cityscapes~\citep{cordts2016cityscapes}.  We follow the same preprocessing protocol as adopted in
PiCIE~\citep{cho2021picie} and STEGO~\citep{hamilton2022unsupervised}: images are first resized so the minor edge is
320px and then cropped in the center to produce square images.

\subsection{Optimization}

In the per-image setting we choose each head to be an independent affinity graph, but that leads to extremely
expensive experiments at the full-dataset level.  To control this expense, we experiment with a few alternatives:
considering each head independently and sampling random layers and heads per iteration of optimization, or
concatenating the features for each head into one large vector, which reduces the number of graphs by a factor of 8.
The second ultimately led to better results.  Due to prohibitive memory costs, we also only consider attention layers
with resolutions of 32x32 or coarser.  This avoids the large graphs constructed by layers with 64x64 resolution.  Due
to the prohibitive cost associated with optimizing one set of features per image in the dataset, we restrict our
dataset-level clustering to the validation set only.

\subsection{Evaluation}

\textbf{Unsupervised semantic segmentation.}
We consider $\mX_\text{ortho}$ as features for our method.  We also compare with several baseline methods by collecting
backbone features from a number of different models: STEGO, DINO and Stable Diffusion.  For Stable Diffusion we choose
the most semantic features in the model, as measured by semantic correspondence performance in prior
work~\citep{tang2023emergent}.  For DINO we take features at the last layer, like prior work~\cite{wang2022tokencut,
wang2023cut,hamilton2022unsupervised}.  For STEGO, we use output just before the linear head that projects to the
number of clusters.

After obtaining features, we cluster with $K = 27$, the number of ground truth categories in both datasets, for K-Means
over $\mX_\text{ortho}$.  We report results with mIoU and compare to other methods in Table~\ref{tab:kmeans_coco}.

\vspace{0.25em}
\noindent
\textbf{X-Y coordinate regression.}
After extracting features, we use a random sample of 80\% of the features to learn a linear regression model onto the
X-Y coordinates of a 32 x 32 grid, and check performance on the remaining 20\%.  For methods where the features are of
a different resolution, we resize bilinearly.

\section{More Applications of Per-Image Regions}

\subsection{Adapting CLIP for Open-Vocabulary Semantic Segmentation}\label{app:clip_open_vocab_seg}

\begin{figure}[t]
   \begin{center}
      \includegraphics[width=1.0\linewidth]{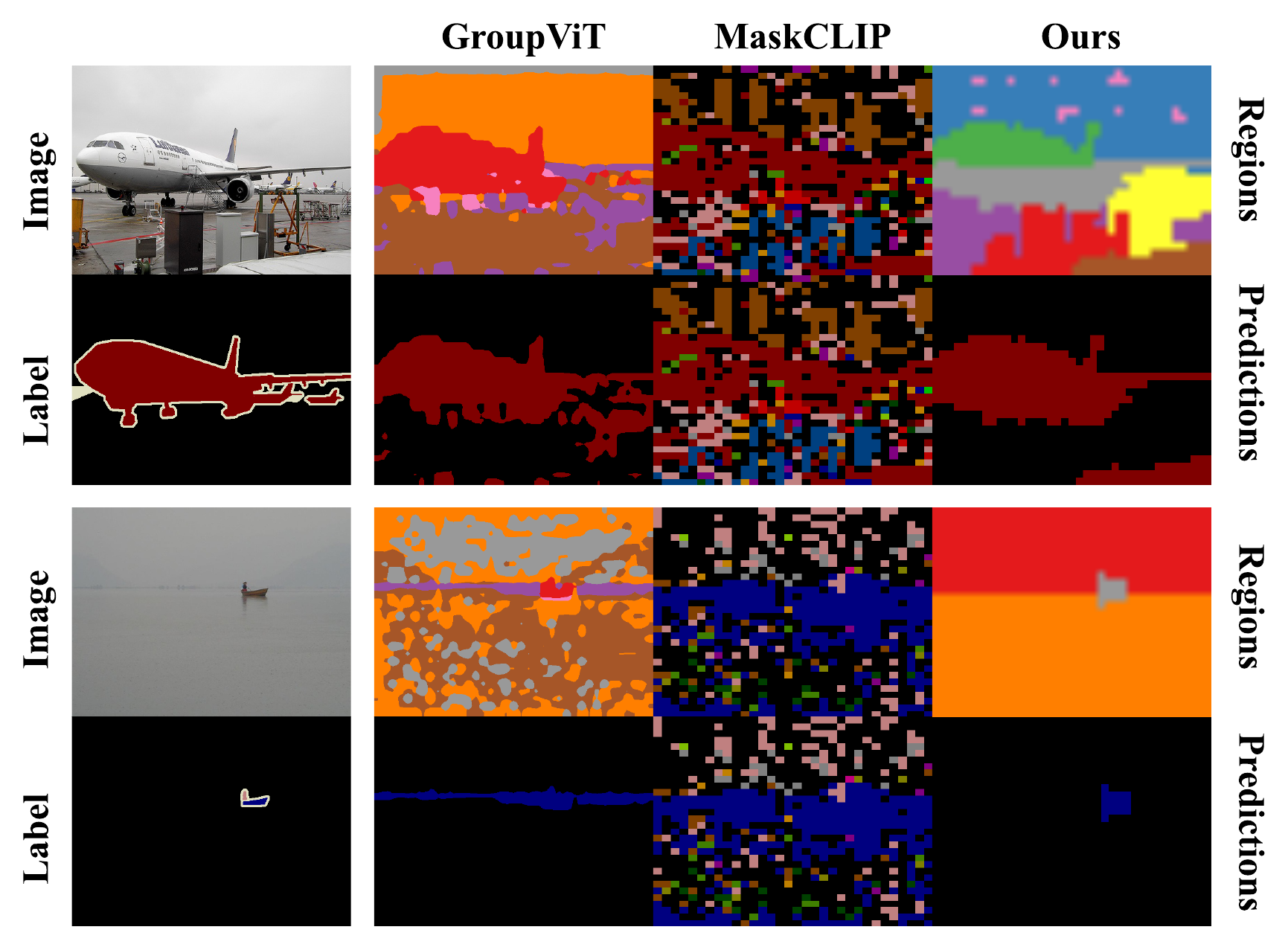}
   \end{center}
   \vspace{-1.0em}
   \caption{%
      \textbf{Examples for different methods on zero-shot semantic segmentation.}
      Notice the tendency of GroupViT~\cite{xu2022groupvit} and MaskCLIP~\cite{zhou2022extract} to break up objects,
      and the eagerness of MaskCLIP to cover the image.  On the airplane image we perform slightly worse than GroupViT
      but our regions have more spatially coherent structures.  On the boat image our method has better performance
      and can even separate water and sky, though the gap between their pixel values is almost imperceptible.%
   }%
   \label{fig:zero_shot}
\end{figure}

As a more interesting case-study than oracle decoding, we assess our regions for zero-shot semantic segmentation on
PASCAL VOC~\cite{everingham2015pascal}.  In order to form class decisions, we follow insights from
GroupViT~\cite{xu2022groupvit} and MaskCLIP~\cite{zhou2022extract}.  First we compute regions on top of CLIP ViT-B/16,
then we take the final value vectors from the last attention layer as pixel-wise features, similar to MaskCLIP.  We
compute region-wise features by averaging pixel-wise features over the regions they correspond to, then compute cosine
similarities between these region-wise features and the text embeddings of CLIP, where per-class text embeddings are
computed by an average over many different prompts like \textit{``a photo of a \{class name\}, a picture of a
\{class name\}, ...''}, as is done in GroupViT.  Finally we threshold these similarities by a fixed number (0.7), and
set all regions to their most similar class, where regions with no similarity greater than the threshold are assigned
background.  We compare to MaskCLIP~\cite{zhou2022extract}, a training-free approach, as well as
GroupViT~\cite{xu2022groupvit}, which proposes modifications to the original CLIP architecture in order to better suit
segmentation.

\begin{table}
   \begin{center}
   \footnotesize
   \setlength{\tabcolsep}{4.75pt}
   \begin{tabular}{@{}l|c|c|c@{}}
      \bf Method & \bf Segmentation-specific? & \bf Model & \bf mIoU \\
      \HLINE
      GroupViT \cite{xu2022groupvit} & Yes & modified ViT-S/16 & 0.53 \\
      \hline
      MaskCLIP \cite{zhou2022extract} & No & ViT-B/16 & 0.25 \\
      Ours & No & ViT-B/16 & 0.50
   \end{tabular}
   \end{center}
   \vspace{-1.0em}
   \caption{%
      \textbf{Zero-shot segmentation on PASCAL VOC \cite{everingham2015pascal}.}
      Our method is stronger than the MaskCLIP baseline, and competitive with GroupViT, whose architecture is tailored
      to segmentation.%
   }%
   \label{table:seg_specific}
\end{table}

We see in Table~\ref{table:seg_specific} that, even without a segmentation-specific training objective, we can achieve
competitive performance on PASCAL VOC~\cite{everingham2015pascal}, and our region-extraction pipeline aids in
segmentation on top of CLIP~\cite{radford2021learning}.  We emphasize that this is possible
\textit{without any segmentation-specific objectives or additional training}.

\begin{figure}[t]
   \begin{center}
      \includegraphics[width=1.0\linewidth]{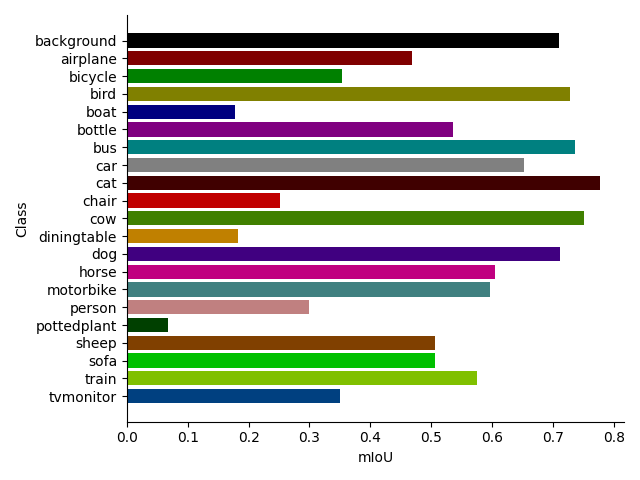}
   \end{center}
   \vspace{-1.0em}
   \caption{%
      \textbf{Per-class mIoUs on PASCAL VOC.}
      Errors are pronounced in a few particular classes, like ``boat'', ``potted plant'', and ``dining table,''
      which are primarily due to localization issues with CLIP.%
   }%
\label{fig:per_class}
\end{figure}

\begin{figure}[t]
   \begin{center}
      \includegraphics[width=1.0\linewidth]{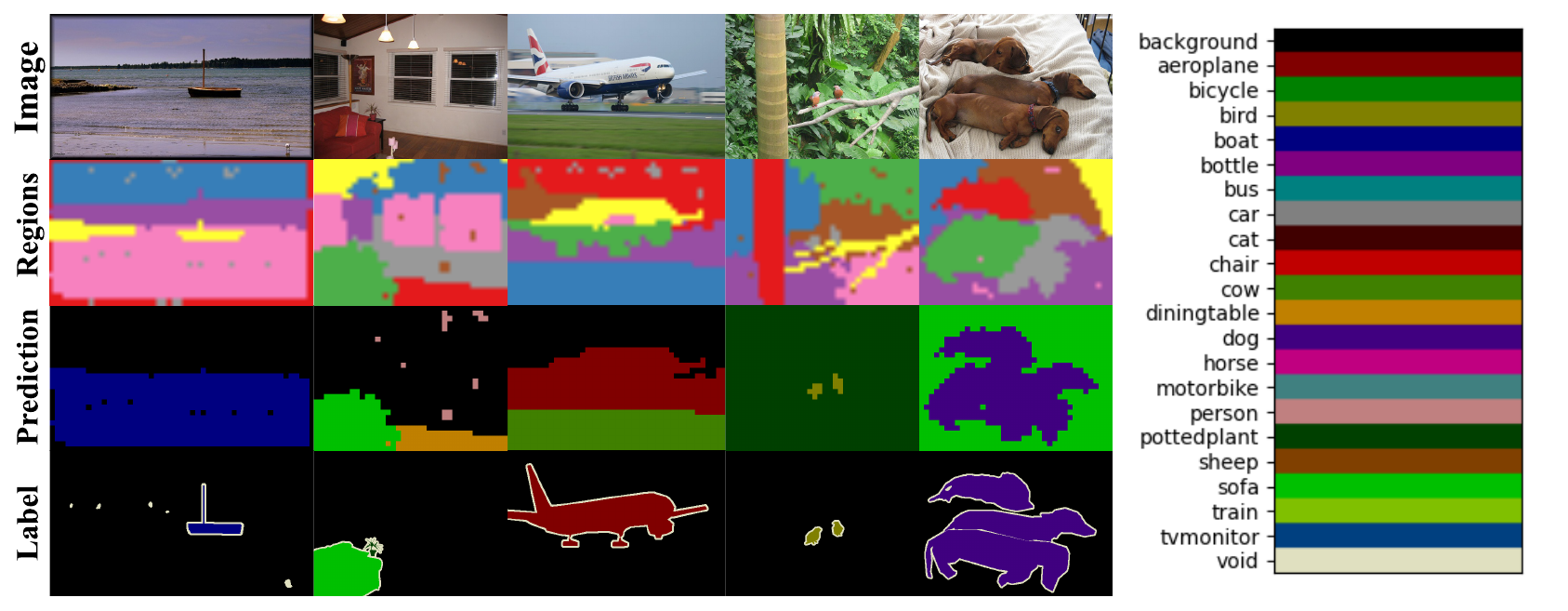}
   \end{center}
   \vspace{-1.0em}
   \caption{%
      \textbf{Examples of segmentation failures.}%
      From the regions we see that most object are correctly segmented and classified, but CLIP fails on the
      background.  From left to right, water is classified as ``boat,'' hardwood floor is classified as
      ``dining table,'' runway grass is classified as ``cow,'' forest foliage is classified as ``potted plant,''
      and bedding is classified as ``sofa.''  This persists across threshold values, as the CLIP similarities are
      very high.  Refining CLIP's localization ability can close much of the gap to oracle decoding.%
   }%
   \label{fig:seg_failures}
\end{figure}

Our regions are often contiguous and large in size, while GroupViT's regions contain holes.  As a result, the errors
that CLIP makes in localizing certain classes may be magnified by our regions.  This can be seen in per-class IoU
scores in Figure~\ref{fig:per_class}, and examples of CLIP's failure to localize in Figure~\ref{fig:seg_failures}.
Crucially, it appears that CLIP does a poor job localizing particular classes, associating ``boat'' to any water or
beach in the image, ``potted plant'' and ``cow'' to ground cover, and ``person'' to all sorts of human-built objects.
Fixing these localization errors in CLIP is out of the scope of our contributions, but could yield improvements to
match segmentation-specific methods.

\subsection{Unsupervised Instance Segmentation}\label{app:instance_seg}

As an additional proof-of-concept, we run experiments on a more difficult task, unsupervised instance segmentation,
which requires simultaneously generating object proposals and segmenting salient objects.  To benchmark our method,
we use the standard COCO 2017~\citep{lin2014microsoft} validation split, and follow prior work~\citep{wang2023cut} to
report results on both instance segmentation and object detection metrics in a class-agnostic setting.  Due to the
difficulty of generating instance proposals in a diverse image distribution, recent attempts~\citep{wang2022freesolo,
wang2022tokencut} design heuristic decoding strategies based on the structure of a particular model's features,
\eg~the final layer of DINO~\citep{caron2021emerging}, in order to generate region proposals.

However, we hypothesize that, if the features are informative enough, a simple clustering strategy and generic scoring
function should suffice for high-quality instance segmentation.  In our implementation, we use K-Means to generate
region proposals, and silhouette scores to rank those proposals.

We start by generating initial region proposals by clustering with K-Means on top of the dense features we extract,
with $K$ ranging from 2 to 10.  To further expand our pool of proposals, we use agglomerative clustering to
hierarchically merge spatially adjacent regions with ward linkage.

Naively, we can treat each instance proposal as a binary clustering problem with the foreground and background each as
their own cluster, and directly use silhouette scores to rank proposals.  However, instances usually take up a
relatively small portion of an image making the binary clustering extremely imbalanced, which significantly harms the
scores and ranking.

To this end, instead of treating the complement of foreground masks as background, we subsample the background pixels
to create a balanced subset by only preserving the background pixels that are close to the foreground pixels in feature
space.  We also adopt standard post-processing steps to remove duplicate and extreme-sized segments before producing
the final output.  Finally, since the silhouette score is in the range $[-1,1]$, we can use 0 as a threshold to remove
low-quality proposals.

We follow the above procedure on top of the features produced by optimizing Eqn.~\ref{eqn:primary} over Stable
Diffusion's attention layers.  We report our results and compare to the current state-of-the-art region proposal
methods in Table~\ref{tab:instance_segmentation}.

\begin{figure}[t]
   \begin{center}
      \includegraphics[width=1.0\linewidth]{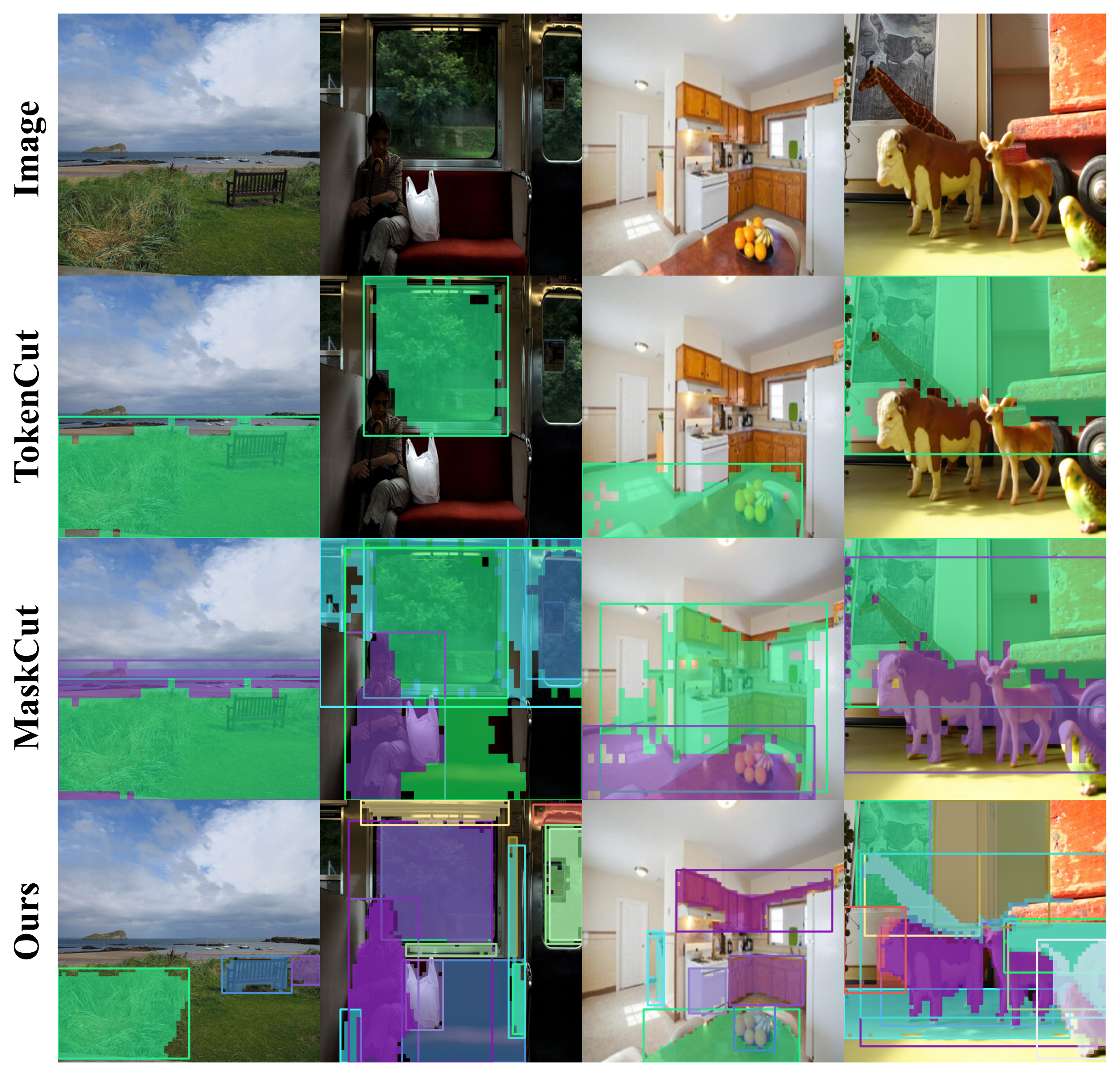}
   \end{center}
   \vspace{-1.0em}
   \caption{%
      \textbf{Examples of different methods on instance segmentation.}
      As described by the original authors, TokenCut~\cite{wang2022tokencut} can only generate a single object
      proposal, and MaskCut~\cite{wang2023cut} is limited as well.  Our method shows better localization results and
      scales to many instances.%
   }%
   \label{fig:instance_segmentation}
\end{figure}

Due to the approximation error in binarizing the affinity matrix for clustering, both TokenCut and MaskCut have
trouble yielding diversified samples.  By contrast, our learned features contain richer information that allows us to
adopt a generic instance grouping pipeline without any post-processing on the features.  As we see in
Table~\ref{tab:instance_segmentation}, this leads us to generate high-quality diversified proposals with better recall
in both instance segmentation and object detection metrics, while maintaining comparable precision to prior methods.
Qualitative results are available in Figure~\ref{fig:instance_segmentation}.

\begin{table}
   \begin{center}
   \footnotesize
   \setlength{\tabcolsep}{1.375pt}
   \begin{tabular}{@{}lcccccccc@{}}
      \bf Method & \#Masks & $AP_{50}^{\text{box}}$ & $AP^{\text{box}}$ & $AR_{100}^{\text{box}}$&$AP_{50}^{\text{mask}}$ & $AP_{\text{mask}}$ & $AR_{100}^{\text{mask}}$\\
      \HLINE
      TokenCut \cite{wang2022tokencut} &1 & 5.2 & 2.6 & 5.0 & 4.9 & 2.0 & 4.4\\
      TokenCut \cite{wang2022tokencut} &3  & 4.7 & 1.7 & 8.1 & 3.6 & 1.2 & 6.9\\
      MaskCut \cite{wang2023cut} &3 & 6.0 & 2.9 & 8.1 & 4.9 & 2.2 & 6.9\\
      \hline
      Ours &13 & 4.0 & 1.9 & \textbf{11.2} & 4.0 & 1.5 & \textbf{8.2}\\
   \end{tabular}
   \end{center}
   \vspace{-1.0em}
   \caption{
      \textbf{Results of instance segmentation on COCO-val-2017~\cite{lin2014microsoft}.}
      Our learned pixel-wise features, with a simple and generic instance segmentation decoding pipeline,
      significantly outperform baselines in recall, in both object detection and instance segmentation.  On the other
      hand, despite generating many more proposals per image, our method still maintains comparable precision.%
   }%
   \label{tab:instance_segmentation}
\end{table}

\section{Code Sources}

All experiments are implemented in Python with PyTorch~\cite{paszke2019pytorch}.  For Stable
Diffusion~\cite{rombach2022high}, we use HuggingFace Diffusers~\cite{von-platen-etal-2022-diffusers}.  For baselines,
we use official numbers, implementations, and model weights, except in the case of MaskCLIP~\cite{zhou2022extract},
where we reimplement the method due to difficulty in obtaining satisfactory performance.

\end{document}